\def\year{2022}\relax
\documentclass[letterpaper]{article} 
\usepackage{amsmath,amsfonts,amssymb,bm}
\usepackage{bbm}
\usepackage{aaai22}  
\usepackage{times}  
\usepackage{helvet}  
\usepackage{courier}  
\usepackage[hyphens]{url}  
\usepackage{graphicx} 
\usepackage{natbib}  
\usepackage{caption} 
\DeclareCaptionStyle{ruled}{labelfont=normalfont,labelsep=colon,strut=off} 
\usepackage{algorithm,algpseudocode}
\usepackage{algorithmicx}
\usepackage{algpseudocode} 
\usepackage{dcolumn}
\usepackage{booktabs}
\usepackage{multirow}
\usepackage{graphicx}
\usepackage{subfigure}
\usepackage{color}
\usepackage{newfloat}
\usepackage{listings}
\floatstyle{ruled}
\newfloat{listing}{tb}{lst}{}
\floatname{listing}{Listing}
\usepackage{enumerate}

\nocopyright 

\title{A Closer Look at Personalization in Federated Image Classification}
\author {
    Changxing Jing,\textsuperscript{\rm 1}
    Yan Huang, \textsuperscript{\rm 2}
    Yihong Zhuang \textsuperscript{\rm 1}
    Liyan Sun \textsuperscript{\rm 1}
    Yue Huang \textsuperscript{\rm 1}
    Zhenlong Xiao \textsuperscript{\rm 1}
    Xinghao Ding \textsuperscript{\rm 1} \thanks{Corresponding author: dxh@xmu.edu.cn}
}
\affiliations {
    \textsuperscript{\rm 1} School of Informatics, Xiamen University, Xiamen, Fujian, 361005, China\\
    \textsuperscript{\rm 2} College of Computing and Software Engineering, Kennesaw State University, Kennesaw, GA, 30144, U.S.A.\\
    
}

\usepackage{bibentry}

\begin{document}

\maketitle

\begin{abstract}
Federated Learning (FL) is developed to learn a single global model across the decentralized data, while is susceptible when realizing client-specific personalization in the presence of statistical heterogeneity. 
However, studies focus on learning a robust global model or personalized classifiers, which yield divergence due to inconsistent objectives. 
This paper shows that it is possible to achieve flexible personalization after the convergence of the global model by introducing representation learning. 
In this paper, we first analyze and determine that non-IID data harms representation learning of the global model. 
Existing FL methods adhere to the scheme of jointly learning representations and classifiers, where the global model is an average of classification-based local models that are consistently subject to heterogeneity from non-IID data.
As a solution, we separate representation learning from classification learning in FL and propose \texttt{RepPer}, an independent two-stage personalized FL framework.
We first learn the client-side feature representation models that are robust to non-IID data and aggregate them into a global common representation model.
After that, we achieve personalization by learning a classifier head for each client, based on the common representation obtained at the former stage.
Notably, the proposed two-stage learning scheme of \texttt{RepPer} can be potentially used for lightweight edge computing that involves devices with constrained computation power.
Experiments on various datasets (CIFAR-10/100, CINIC-10) and heterogeneous data setup show that \texttt{RepPer} outperforms alternatives in flexibility and personalization on non-IID data.
\end{abstract}

\section{Introduction}
The enormous amount of edge devices and various terminals continually generating large-scale datasets, which draw a magnificent concern regarding data privacy and sensitivity~\cite{feddata-1,feddata-2,feddata-3}.  
Federated learning (FL)~\cite{fedavg}, a distributed machine learning paradigm, has shown great promise in reducing privacy risk and communication costs~\cite{advance,challenge,survey1}. 
It enables multiple clients to learn a global model collaboratively over distributed partitions of data under the management of a central server with a built-in privacy-preserving design.
One challenge associated with decentralized data in FL is statistically heterogeneous across the client~\cite{divergence-1,divergence-2,zhang2021federated}. 
The non-IID data due to the different contexts and preferences of the distributed clients.
Indeed, statistical heterogeneity describes distributions of labels among clients.
Specifically, local computations in a supervised setting on such non-IID data are unavoidably inherited with label bias, bringing local models to drift significantly from each other~\cite{FRT:FedProx,scaffold,RN43}.
Consequently, the resultant global model may result in poor decentralized data adaptation. 

\begin{figure}[ht]
	\centering
	\subfigure[FedAvg]{\includegraphics[width=0.42\columnwidth]{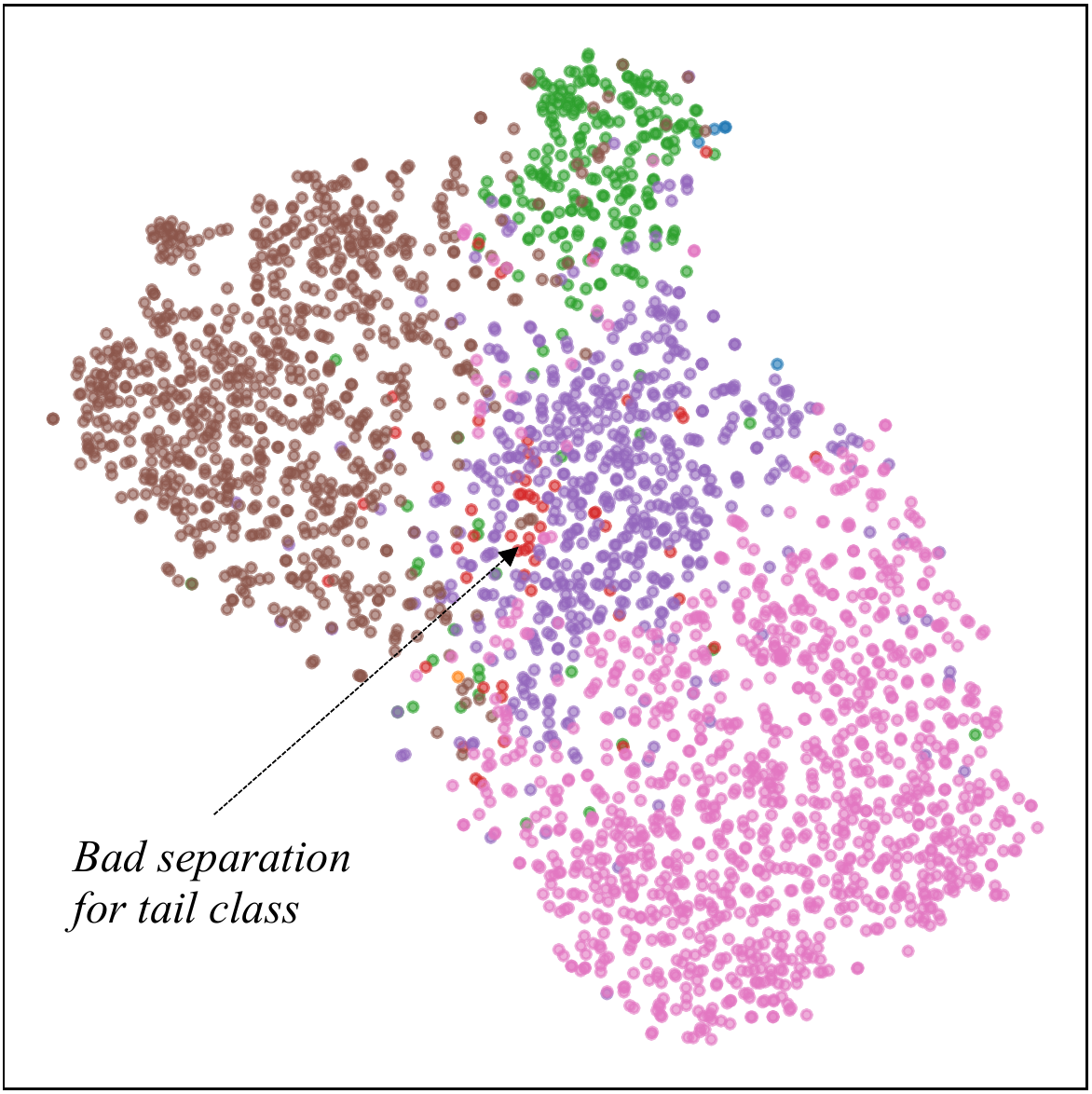}\label{imb:fedavg}} 
	\subfigure[LG-FedAvg]{\includegraphics[width=0.42\columnwidth]{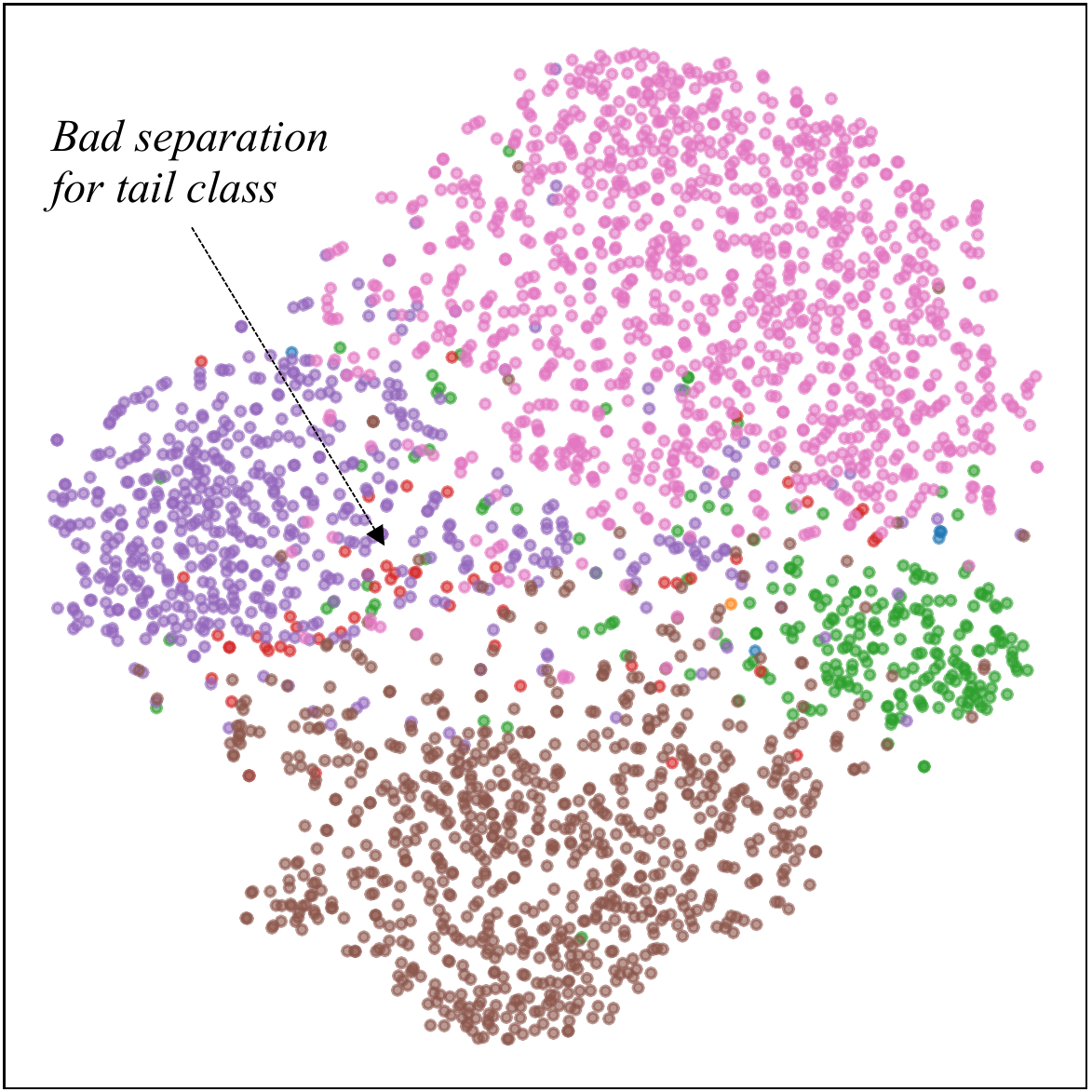}\label{imb:lg}} 
	\subfigure[FedRep]{\includegraphics[width=0.42\columnwidth]{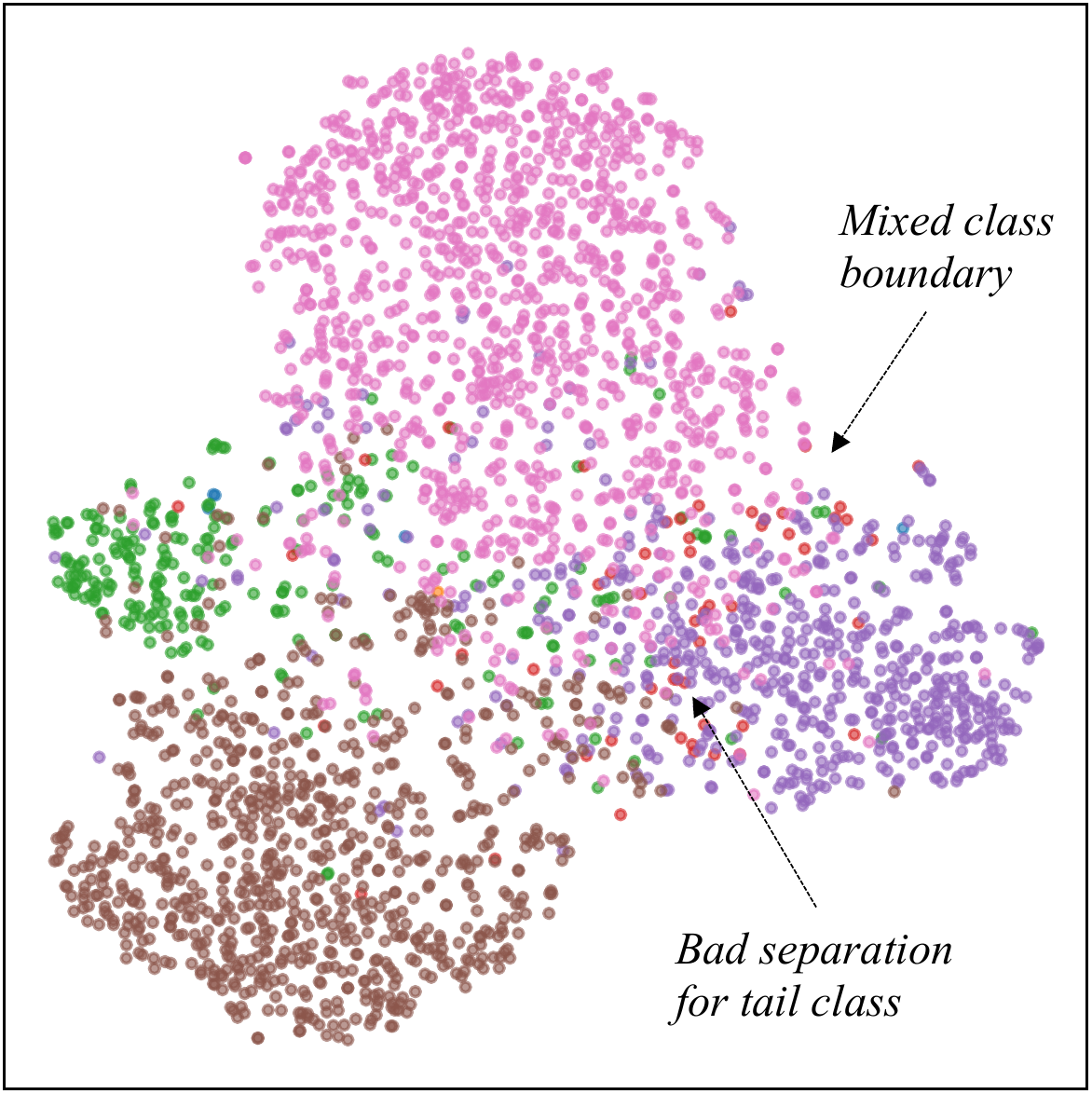}\label{imb:fedrep}} 
	\subfigure[{\bf \texttt{RepPer}}]{\includegraphics[width=0.42\columnwidth]{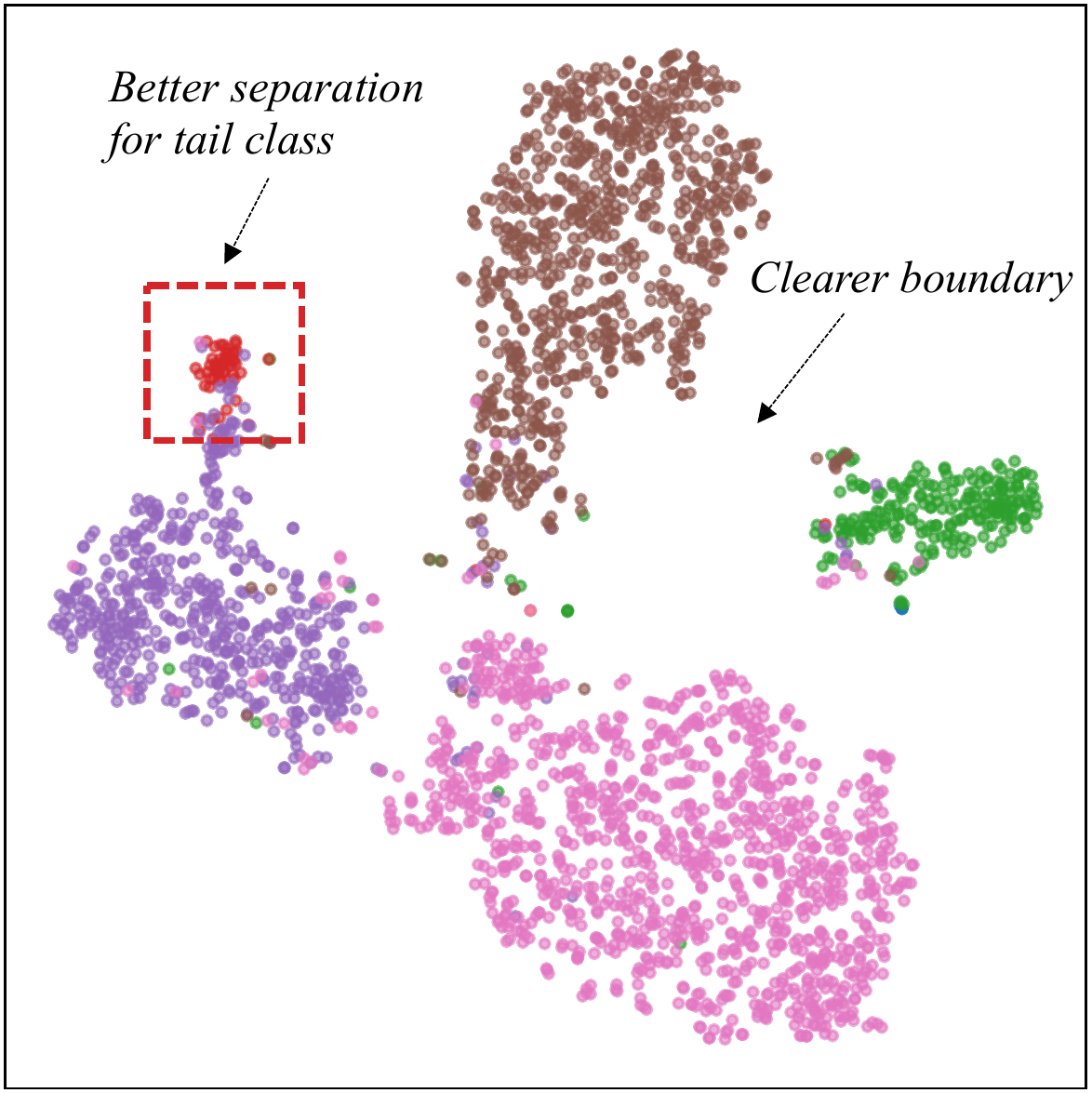}\label{imb:repper}} 
	\caption{Comparing the feature representations of \texttt{RepPer} with the jointly learning schemes, including FedAvg~\cite{fedavg}, LG-FedAvg~\cite{Rep:LG-Fedavg}, and FedRep~\cite{Rep:FedRep} on the local data of the CIFAR-10 dataset under the non-IID settings. 
		TSNE visualization shows that non-IID data can result in poor class separation, particularly for instance-scarce (or tail) classes. 
		The vanilla FL and FRL approaches suffer from biased and mixed representations, while \texttt{RepPer} benefits to shape clear class boundaries, leading to more apparent separation and improved performance.
		The classification accuracy comparison results are shown in Table.~\ref{table:comparison}.
	}
	\label{fig:imb}
\end{figure}

The problem of statistical heterogeneity in FL has been extensively investigated in the literature on personalized FL.  
Two active strands of work involve learning a robust global model with meta-learning~\cite{meta:maml,meta:fedmeta,meta:aruba,meta:per-fedavg} and regularization training for better generalization performance~\cite{RT:pFedMe,RT:FedAMP}, and personalized models on heterogeneous data with transfer learning~\cite{trans:FedPer,trans:FedHealth}. 	
Recent advances fueled by contrastive representation learning~\cite{ssl:cmc,ssl:moco,ssl:byol,ssl:simclr,ssl:simsam} suggest a possible route for further improvements of FL. 
Specifically, federated representation learning (FRL) suggests that learning local representations among their local data and exploring the reduced-dimensional representations on the server can alleviate the influence of statistic heterogeneous on personalized FL~\cite{Rep:FedCA,Rep:BYOL,bridging,fedbabu}.
Though the FRL methods are intuitive and experimentally effective, they are not without limitations: e.g., the local representations need to be uploaded to the server, which results in an increased risk of privacy leakage.
More importantly, representation and classification training are \textit{bound together} in a supervised fashion (even if the decoupling process is subsequently performed in FedRep~\cite{Rep:FedRep}). 
The local training is driven by the client's empirical risk, which will inevitably be affected by label bias~\cite{non-iid:cls}. 
We further observe that non-IID data lead to biased feature representations.
This is depicted in Figure~\ref{fig:imb}, where the feature representations of the global model of the FedAvg~\cite{fedavg}, LG-FedAvg~\cite{Rep:LG-Fedavg}, FedRep~\cite{Rep:FedRep}, and \texttt{RepPer} on local data. 
This observation illustrates that the negative effects of non-IID persist not only in the classification accuracy but also in the feature representations. 

\textit{Is it possible to learn and exploit a common representation model on non-IID data while maintaining good personalized predictions for all the clients?} 
To answer this question affirmatively, we propose \texttt{RepPer}, a two-stage framework for learning from non-IID data, which eliminates the correlation between representation learning and personalized prediction.
Instead of learning representations from the classification objective, our key idea is to separate representation learning from classification in the local update, from which to mitigate the adverse effect of label bias and client-drift.
Specifically, we built a two-stage training procedure:
(1) Common representation learning (CRL) stage. 
We construct a common feature space with discriminative capability by averaging local representation model updates.
The participants proceed with local computations based on supervised contrastive (SC) loss~\cite{supcon} to learn local representation models and update the global representation model similar to the standard FedAvg. 
(2) Personalized classification learning (PCL) stage.
We learn a personalized classifier head for each client on their distribution, using the common representation model obtained from the CRL stage. 
Since the two stages of \texttt{RepPer} are optimized independently and separately, every client can obtain a fully customized classification model with low complexity after the first stage has been converged.
As shown in Fig.~\ref{figz}, we show that the parameters of the global representation model can be updated by aggregating local representation model parameters, where each local model aims to learn feature representation from the heterogeneous data. 
Local personalization depends on satisfying client-specific target distributions upon the learned global representation model. 
With this strategy, local clients learn discriminative feature representation in the CRL stage and leverage the aggregated global representation model to optimize their personalized classifiers w.r.t local data in the PCL stage. 

\texttt{RepPer}  is flexible in Internet of Things (IoT) applications in the actual federated scenario, especially in conditions where edge devices have less computation power.
In \texttt{RepPer}, personalized classifiers can be flexibly designed by traditional machine learning and deep learning techniques, such as support vector machine (SVM~\cite{svm}), logistic regression (LR) and multi-layer perceptron (MLP) neural network.
Even for clients who can only support classification that is incompetent to participate in feature representation, they can flexibly train their personalized classifier derived from the common representation from the CRL stage. 

We evaluate \texttt{RepPer} on federated image classification and show that it outperforms recently proposed alternatives on different levels of statistical heterogeneity among clients. 
We also consider and perform experiments on out-of-local-distribution generalization, wherein one client personalization can optimize for newly targeting distributions in the federation that differ from the raw data distribution. 
Finally, we explore the flexibility for specific clients considering the insufficient computational power in the realistic federation.

The main contributions of this work can be summarized as follows:
\begin{itemize}
	\item [i)]
	{We establish  \texttt{RepPer}, an independent two-stage personalized FL framework that separates traditional FL into representation and classification learning. 
		First, we explore to learn a common representation model from the non-IID data. 
		Then, each client can design a personalized classifier on their local data flexibly by using the well-learned common representation model.}
	\item [ii)]
	{ We make a practical consideration of FL in edge computing and out-of-local-distribution generalization. 
		The \texttt{RepPer} allows edge devices with different computing powers to participate in FL. 
		New clients or target distributions can be well generalized based on \texttt{RepPer} when the CRL stage is available.} 
	\item [iii)]
	Experimental results validate the generalization and classification accuracy of the \texttt{RepPer} in 
	(a) different levels of statistical heterogeneity; 
	(b) generalization on out-of-local-distribution data; 
	(c) various computing powers devices.
\end{itemize}

\section{Related Work}
\subsection{Federated Learning} 
In FL, the central server coordinating a total of \textit{K} clients jointly solve
the following optimization problem:
\begin{equation}
	\begin{aligned}
		\underset {w}{\min}\left\{F(w):=\sum_{i=1}^{K}q_i f_i(w)\right\},
		\label{fedavg}
	\end{aligned}
\end{equation}
where the global objective function $F(w)$ is the average of the local objectives $f_i(w)$ with the weight $q_i$ of the participant \textit{K} clients. 
In particular, $f_i(w)$ measures local empirical risk across local data distribution $D_i=\{x_1^i,x_2^i,\cdots,x_{n_i}^i\}$, defined as
\begin{equation}
	f_i(w)=\frac{1}{n_i}\sum_{j=1}^{n_i}f_i(w; x_j^i),
	\label{eq:f_i}
\end{equation} 
where $n_i$ is the count of indices of samples on \textit{i}-th client, $n=\sum_i(n_i)$ is the sum of samples across all the clients. 
We set $q_i={n_i}/{n}$, where $i\in [K]$. 
Recent methods have studied the personalization of FL over multiple sources of non-IID data. 
The personalized FL optimizes the objective in Eq.~\eqref{fedavg} and~\eqref{eq:f_i}, aiming to 1) learn a global model $w$ from the decentralized data; 2) achieve client-specific personalization. 
Federated meta-learning proposes to find a good initial condition shared across participating clients as an initial global model and then optimize for personalization in cooperating with meta-learning~\cite{meta:fedmeta,meta:per-fedavg,meta:aruba}.
Federated transfer learning offers to transfer the global model to each client by freezing the distributed lower layers of the global model while fine-turning its higher layers in terms of local data~\cite{trans:FedPer,trans:FedHealth}.
Federated regularization training introduces a unit $L_2$-norm to constrain the difference of model parameters between the global and local models to stabilize convergence~\cite{RT:FedAMP,RT:pFedMe,FRT:FedProx}.
In all of these methods, each client in the federation is trained in a jointly learning scheme, limited by label bias and client-drift from non-IID data.

\subsection{Contrastive Representation Learning} 
Contrastive representation learning has seen remarkable success in learning representations, especially on unlabeled data~\cite{ssl:cmc,ssl:moco,ssl:byol,ssl:simclr,ssl:simsam}.
The common motivation behind it is introducing a contrastive loss~\cite{ssl:nce} in representation learning. 
Contrastive loss maximizes the consistency between augmented views of the same image by contrasting the agreement between different images. 
Supervised contrastive learning~\cite{supcon} incorporates label information to maximize features from the same class.
Contrastive representation learning has been widely investigated in long-tailed and class imbalanced classification~\cite{kang2020exploring,wang2021contrastive}. 
It helps to learn discriminative features and ease classifier learning in imbalance cases.

\subsection{Federated Representation Learning}
Recent researchers focus on learning representations across participant clients on heterogeneous data, and further optimizing personalization for each client.
FedCA~\cite{Rep:FedCA} and FedU~\cite{Rep:BYOL} learn representations from unlabeled non-IID data.
LG-FedAvg~\cite{Rep:LG-Fedavg} and FedRep~\cite{Rep:FedRep} are similar to \texttt{RepPer}, but they jointly learn representation on each local data constrained by classification objectives, such as cross-entropy loss, leading to biased representations, as shown in Fig.~\ref{imb:lg} and~\ref{imb:fedrep}.
Moreover, the most recent FRL methods upload the representations learned from local data, leading to privacy leakage. 
Our approach separates representation learning from classification training in local updating. 
We introduce the recently proposed supervised contrastive (SC) loss for the local representation model updating on non-IID data, and prevent transferring feature representation vectors to the server to avoid privacy leakage. 
\texttt{RepPer} is a general framework that mitigates the non-IID problem, makes flexibility, and reduces personalization computation. 

\begin{figure*}
	\centering
	\includegraphics[width=0.70\textwidth]{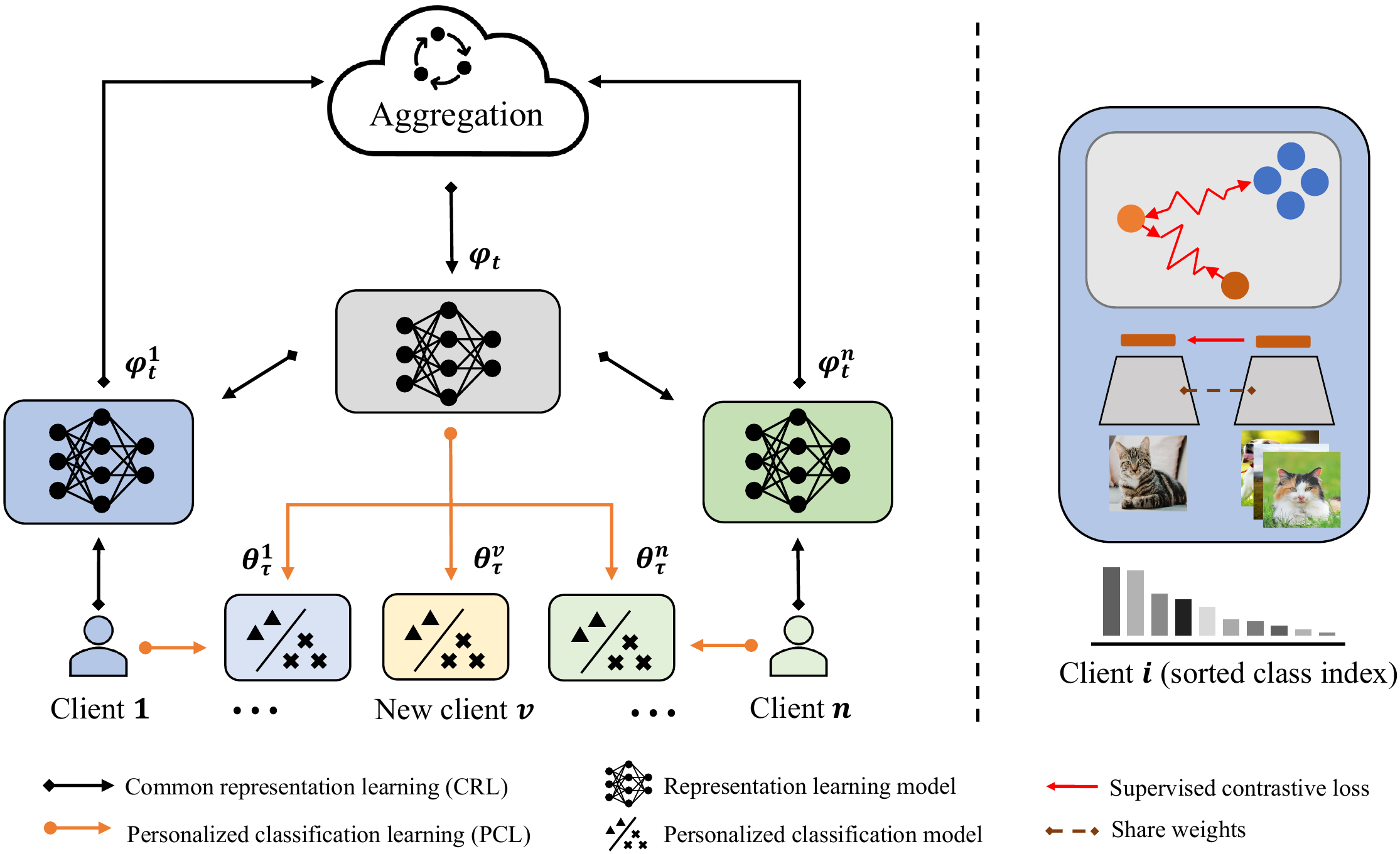}
	\caption{\textbf{Left}: \texttt{RepPer} allows local and global feature representation learning on heterogeneous data and flexible design in personalized classification. 
		Black line: Common representation learning (CRL) stage shows client-server synchronization representation optimization in the federation. 
		Blue and green indicate local representation models, and grey denotes the global server.  
		Each client communicates the local representation model updates to the server. 
		There these updates are aggregated to generate a shared global representation model. 
		Orange line: Personalized classification learning (PCL) stage shows that each local client leverages the resulting global representation model to optimize its desired personalized classifier with local data. 
		Similarly, it is only necessary to update the new classifier head according to the global representation model when facing new client data. 
		\textbf{Right:} Local representation training with the supervised contrastive (SC) loss~\cite{supcon} on client data. 
		The SC loss focuses on clustering the features with semantic discrimination, resulting in less skewed features. }
	\label{figz}
\end{figure*}

\section{Methodology}
To effectively learn an independent common representation model on non-IID data and perform personalized prediction for each client on local data, we show a two-stage personalized federated learning with simple procedures.
An overview of \texttt{RepPer} is shown in Fig.~\ref{figz}. 
\\
\textbf{Stage 1: Common Representation Learning (CRL).} 
Typically, the objective $f_i(w;x)$ in Eq.~\eqref{eq:f_i} is generally a cross-entropy loss corresponding to client classification on non-IID data. 
The feature distribution learned on such decentralized data can be highly skewed (shown in Fig.~\ref{fig:imb}), and the decision boundary can be affected by label bias on classification optimization. 
The CRL stage suggests local models to learn distinguishable feature representations in the federation by introducing contrastive semantic clustering constraints. 
To do this, we build on SC loss for local data in each client, which pulls samples with the same class closer in feature space and pushes samples apart from other classes. 
In each communication round of the CRL stage, local representation models learn corresponding feature representations from both instance-rich (or head) and instance-scarce (or tail) classes.
The global representation model allows client-server synchronously updating in the federation.
It should be emphasized that the server takes a weighted average of the local representation model parameters without extra feature representation vectors learned from each client or raw data to decrease the risk of privacy leakage.
\\
{\bf Stage 2: Personalized Classification Learning (PCL).} 
Benefitting from the CRL stage, a personalized classification model is trained for each client using the generated global representation model to create a personalized classifier head on local data. 
Here, we consider different amounts of available computing power on each client in edge computing and adapt to the personalized classification process by learning local classifier heads with different sizes and complexity. 
Some clients who lack computing power in participating in the iterations of FL can flexibly train their personalized low-dimensional classifier, such as SVM, logistic regression, or neural networks, by using the ready-made global representation model from the CRL stage. 
We describe how the stages of CRL (in Section~\ref{sec:3.1}) and PCL (in Section~\ref{sec:3.2}) operate learning on non-IID data.
We then analyze the advantage of representation learning on non-IID data in \texttt{RepPer}.

\subsection{Common Representation Learning (CRL)} \label{sec:3.1}
Instead of optimizing the objective in Eq.~\eqref{fedavg}, we instruct a server to coordinate local clients to train a common representation model. 
Specifically, each client conducts representation learning with SC loss, communicates local representation model updates to the server, and aggregates these updates to the common representation model.
We rewrite the objective Eq.~\eqref{fedavg} to optimize the following objective:
\begin{equation}
	\underset {\phi}{\min}\left\{F(\phi):=\sum_{i=1}^{K}q_i f_i(\phi)\right\}, \label{loss:CLR}
\end{equation}
where the global objective function $F(\phi)$ is the average of local objectives $f_i(\phi)$ weighted by participants.

In local model training, each client learns to map input sample $x\in \mathbb{R}^d$ to a lower-dimensional feature vector $r\in \mathbb{R}^g$ wherein $d\gg g$, 
which is then normalized to the projection vector $z$ onto the unit hypersphere called feature space. 
Then we adopt SC loss to constrain distances of projection vectors from different classes.
In order to increase the number of samples for each class, especially for tail classes, we make two augmentations to each input image $x$.
Thus, each client $i \in [K]$ obtains sample-label pairs $D_i=\{(x_j^i,y_j^i)\}_{j=1}^{2n_i}$ that consist of data augmentations.
In the client $i$, for each sample $x_j^i$ acts as an anchor,
$A(j)$ is the set of all indices in the client $i$ distinct from $x_j^i$,
$P(j) = \{x_p | y_p = y_j, p \ne j\}$ is the set of indices of samples originating from the same class with $x_j^i$ but does not contain $x_j^i$, 
$N(j)=\{A(j)\backslash P(j)\}$ is the set of indices of samples with different classes than $x_j^i$.
Indices in $P(j)$ are called the positives,
and indices in $N(j)$ are called the negatives.
For the projection vectors normalized into the feature space, the SC loss clusters the positives close to the anchor and separates the anchor from the negatives. 
Eq.~\eqref{loss:CLR1}  and ~\eqref{loss:CLR2} present the details of $f_i(\phi)$ in Eq.~\eqref{loss:CLR} as follows:
\begin{equation}
	f_i(\phi) = \sum_{j=1}^{2n_i} \ell_j, \label{loss:CLR1}
\end{equation}
\begin{equation}
	\ell_j=\!-\log\!\left\{\!\frac{1}{|P(j)|}\! \sum_{p\in P(j)}\!\frac{\exp(\boldsymbol{z_j} \cdot \boldsymbol{z_p} / \tau)}{\sum_{a\in A(j)} \exp(\boldsymbol{z_j} \cdot \boldsymbol{z_a} /\tau)}\!\!\right\}, \label{loss:CLR2}
\end{equation}
where $z$ refers to the normalized representation of input $x$, 
the $\cdot$ symbol denotes the inner product,
$\tau \in \mathbb{R}^{+}$ is a scalar temperature parameter. 
Critically, in CRL stage, the label information is required only for clustering feature representations in the feature space rather than intended to classify data from all clients. 

\subsection{Personalized Classification Learning (PCL)} \label{sec:3.2}
After the global representation model has converged, a personalized classifier using global representation can be a much smaller model with less computation.
In PCL stage, each client create a personalized classifier $\theta$ locally, which flexibly fits their client's local data distribution. 
For each client $i \in [K]$, here we train a conventional classification model by minimizing the loss function $\ell_{cls}:\mathbb{R}^C  \times \mathbb{R}^C \to \mathbb{R}$ between the ground-truth and prediction, i.e., cross-entropy loss. 
The goal of this stage for each client is formulated as follows:
\begin{equation}
	\mathop{\arg\min_{\theta_i}}\frac{1}{D_i}\sum_{j\in[D_i]}\ell_{cls}(\theta_i (\phi(x_j^i)), y_j), \label{loss:PCL}
\end{equation}
where $\phi$ is the fixed global representation model learned from the previous CRL stage. 
Each client $i\in [K]$ trains their classifier $\theta_i$ to map from representations to label space. 
Indeed, we study and experiment on various local classification heads with low complexity, including SVM, logistic regression and MLP neural network.
More details are shown in Section~\ref{sec:felxibility}.

\section{Optimization}
During the optimization, \texttt{RepPer} alternates between the client’s local update and a server update on each communication round until convergence in the CRL stage and then optimizes a personalized classifier for each client in the PCL stage. 
Every client performs $\tau_r$ iterations of SGD to compute a local update in the CRL stage and $\tau_c$ iterations to compute in PCL stage.
The subscript $r$ denotes the representation procedure in both client and server updates, and $c$ indicates personalized classification.
The overall training procedure is shown in Algorithm~\ref{alg:a}.

{\bf Client Update.} 
In communication round $t$, a fraction $C\in (0,1]$ of the total clients is uniformly randomly
selected for local updating. 
In the client update, the selected client $i \in [C \cdot K]$ updates the local representation model $\phi_t^i$ with $\tau_r$ iterations by using gradient descent with respect to its joint data as the following:
\begin{equation}
	\phi^i_{t,\tau_r+1} = \phi_{t,\tau_r}^i-\eta_r \nabla f_i(\phi_{t,\tau_r}^i),
\end{equation}
where $\eta_r$ is the learning rate, 
$\nabla f_i(\phi_{t,\tau}^i)$ denotes one step of stochastic gradient according to Eq.~\eqref{loss:CLR1} using the current local representation model $\phi_t^i$. 
Non-selected clients will keep their previous local model parameters.

{\bf Server Update.}
After iterating the local client updates for $\tau_r$ times in round $t$, the participating clients upload parameters with respect to the recent local representation model to the server for aggregation:
\begin{equation}
	\phi_{t+1}=\sum_{i\in [S_t]}q_i \phi_{t}^i,
\end{equation}
where $S_t=\max (C \cdot K, 1)$ is a client set that is randomly selected with a participation rate of $C$ in communication round $t$,
$q_i$ is the weight of the participant client in set $S_t$.

{\bf Personalized Classifier Update.} 
The parameters of classifiers are updated according to the fixed global representation model.
Each personalized classifier only needs a few iteration to converge.
Client $i \in [K]$ updates the current classifier model as follows:
\begin{equation}
	\theta_{{\tau_c}+1}^i= \theta_{\tau_c}^i - \eta_c \nabla \ell_i(\theta_{\tau_c}^i; D_i), \label{update:cls}
\end{equation}
where $\eta_c$ is the learning rate.
Personalized classifiers for each client on their local data can be simply learned using a linear classifier or a shallow neural network. 		

\begin{algorithm}[tb]
	\caption{\textbf{\texttt{RepPer}}}\label{alg:a}
	\textbf{Parameter:}
	$K$ clients are indexed by $i$;
	participation rate $C$; 
	learning rate $\eta_r$, $\eta_c$; 
	number of iterations $\tau_{r}$, $\tau_{c}$;
	number of communication rounds $T$.
	\colorbox[gray]{0.9}{\textbf{Stage 1: Common representation learning}}\\
	{\bf Server executes:}
	\begin{algorithmic} 
		\State initialized global representation model with weights $\phi_0$
		\For{round $t=0,1,\cdots,T-1$}
		\State $m \gets max(C\cdot K,1)$
		\State $S_t \gets $(random set of $m$ clients)
		\For{each client $i\in S_t$ {\bf in parallel}}
		\State $\phi_{t+1}^i \gets$ Client Update ($i, \phi_{t}^{i}$)
		\EndFor
		\State $\phi_{t+1} \gets \sum_{i\in S_t} \frac{D_i}{D} \phi_{t+1}^i$  
		\EndFor
	\end{algorithmic}
	\textbf{Client Update $(i,\phi^i_t)$:}
	\begin{algorithmic}
		\State $B_r \gets$ (split local data $D_i$ into batch)
		\For{$\tau=0,1,\cdots,\tau_{r}-1$}
		\For{batch $b_r \in B_r$}
		\State $\phi^i_{\tau+1} \gets \phi^i_\tau -\eta_r \nabla f_i(\phi^i_\tau; b_r)$
		\EndFor
		\EndFor
		\State return $\phi^i$ to server
	\end{algorithmic}
	~\\
	\colorbox[gray]{0.9}{\textbf{Stage 2: Personalized classification learning}}\\
	\textbf{Clients execute:}  
	//update each local classifier with the global representation $\phi$ frozen
	\begin{algorithmic}
		\State initialized $i$-th local classification model with weights $\theta^i$
		\State $B_c \gets$ (split local data $D_i$ into batch)
		\For{$\tau=0,1,\cdots,\tau_{c}-1$}
		\For{batch $b_c\in B_c$}
		\State $\theta^i_{{\tau}+1} \gets \theta^i_{\tau} - \eta_c \nabla \ell_i(\theta^i_{\tau},\phi;b_c)$ 
		\EndFor
		\EndFor
	\end{algorithmic}
\end{algorithm}

\section{Analysis}
In this section, we provide an analysis of using SC loss for local representation learning on non-IID data.
As shown in Fig.~\ref{fig:imb}, statistic heterogeneity leads to biased representations across clients.
In non-IID settings, clients often contain tail classes associated with only a few samples, which are susceptible to miss-classified of the majority.
In local model training of CRL stage, the SC loss contributes gradient from minorities are large while those for majority samples are small, leading to a more robust local representation clustering both on head and tail classes.
As shown in the Appendix~\ref{A1}, the gradient for Eq.~\eqref{loss:CLR2} with respect to the normalized representation vector $z_j$ has the following form:
\begin{equation}
	\frac{\partial \ell_j}{\partial z_{j}}
	=\frac{1}{\tau}
	\left\{\sum_{p \in P(j)} z_{p}\left(P_{jp} - X_{jp}\right)
	+\sum_{n \in N(j)} z_{n} P_{jn}\right\}, 
\end{equation}
where

\begin{equation}
	P_{j p} = \frac{e^{\left(z_{j} \cdot z_{p} / \tau\right)}}
	{\sum_{a \in A(j)} e^{\left(z_{j} \cdot z_{a} / \tau\right)}},
\end{equation}
\begin{equation}
	X_{j p} = \frac{e^{\left(z_{j} \cdot z_{p} / \tau\right)}}
	{\sum_{p^{\prime} \in P(j)} e{\left(z_{j} \cdot z_{p^{\prime}} / \tau\right)}}.
\end{equation}

In Sec.~\ref{sec:3.1}, we define $r$ the feature vector before normalization, i.e., $z_j=r_j/\|r_j\|$.
The gradient of SC loss with respect to $r$ has the form:
\begin{equation}
	\frac{\partial \ell_{j}}{\partial r_{j}} 
	=	\left.\frac{\partial \ell_{j}}
	{\partial z_{j}}\right|_{\mathrm{P}(\mathrm{j})}
	+\left.\frac{\partial \ell_{j}}
	{\partial z_{j}}\right|_{\mathrm{N}(\mathrm{j})},
\end{equation}
where
\begin{flalign}
	&\ \!\!\frac{\partial \ell_{j}}
	{\partial z_{j}}\Bigg|_{\mathrm{P}(\mathrm{j})}
	\!\!=\frac{1}{\tau \parallel \!\! r_{j} \!\! \parallel}
	\!\! \sum_{p \in P(j)}\!\!\! (z_{p} \!-\! (z_{j}\! \cdot \! z_{p}) z_{j})  (P_{j p} \! -  \!X_{j p}), &\\
	&\ \frac{\partial \ell_{j}}
	{\partial z_{j}}\Bigg|_{\mathrm{N}(\mathrm{j})}
	\!\!=\sum_{n \in N(j)} \left(z_{n} 
	- (z_{j} \cdot z_{n}\right) z_{j}) P_{jn}. &
\end{flalign}	
We show that the SC loss is structured so that tail classes with few samples have large gradient contributions while head classes have small ones. 
If $x_j$ is sampled from the tail class, which acts as an anchor, the set of positives $P(j)$ is irregular and difficult to discriminate from large amounts of negatives, $z_j \cdot z_p \approx 0$, so
\begin{equation}
	\parallel
	\left(z_{p} - \left(z_{j} \cdot z_{p}\right) z_{j}\right)
	\parallel
	=\sqrt{1-\left(z_j \cdot z_p\right)^{2}}
	\approx 1.
\end{equation}
If $x_j$ is sampled from the head class as an anchor, the corresponding positives are numerous, and easy to measure their similarities, $z_j \cdot z_p=1$, thus
\begin{equation}
	\parallel
	\left(z_{p} - \left(z_{j} \cdot z_{p}\right) z_{j}\right)
	\parallel
	=\sqrt{1-\left(z_j \cdot z_p\right)^{2}}
	=0.
\end{equation}

We observe that tail classes with few samples have large gradient contributions, therefore obtaining more compact representations in feature space.
This implicit property avoids representation learning being affected by label bias. 
Furthermore, the label information here applies to robust representation clustering rather than classification tasks. 
Consequently, well-learned feature clustering in each client is the original drive of \texttt{RepPer}.
We provide a full derivation of the property from the gradient descent in Appendix~\ref{A1}.
The common global representation model receives and averages these local updates can perform better clustering than recently proposed methods.
Fig.~\ref{fig:tsne} validates the feature representation capability of the server in \texttt{RepPer}.

\begin{figure*}
	\centering
	\subfigure[CIFAR-10: $\alpha=100$]{\includegraphics[width=0.3\linewidth]{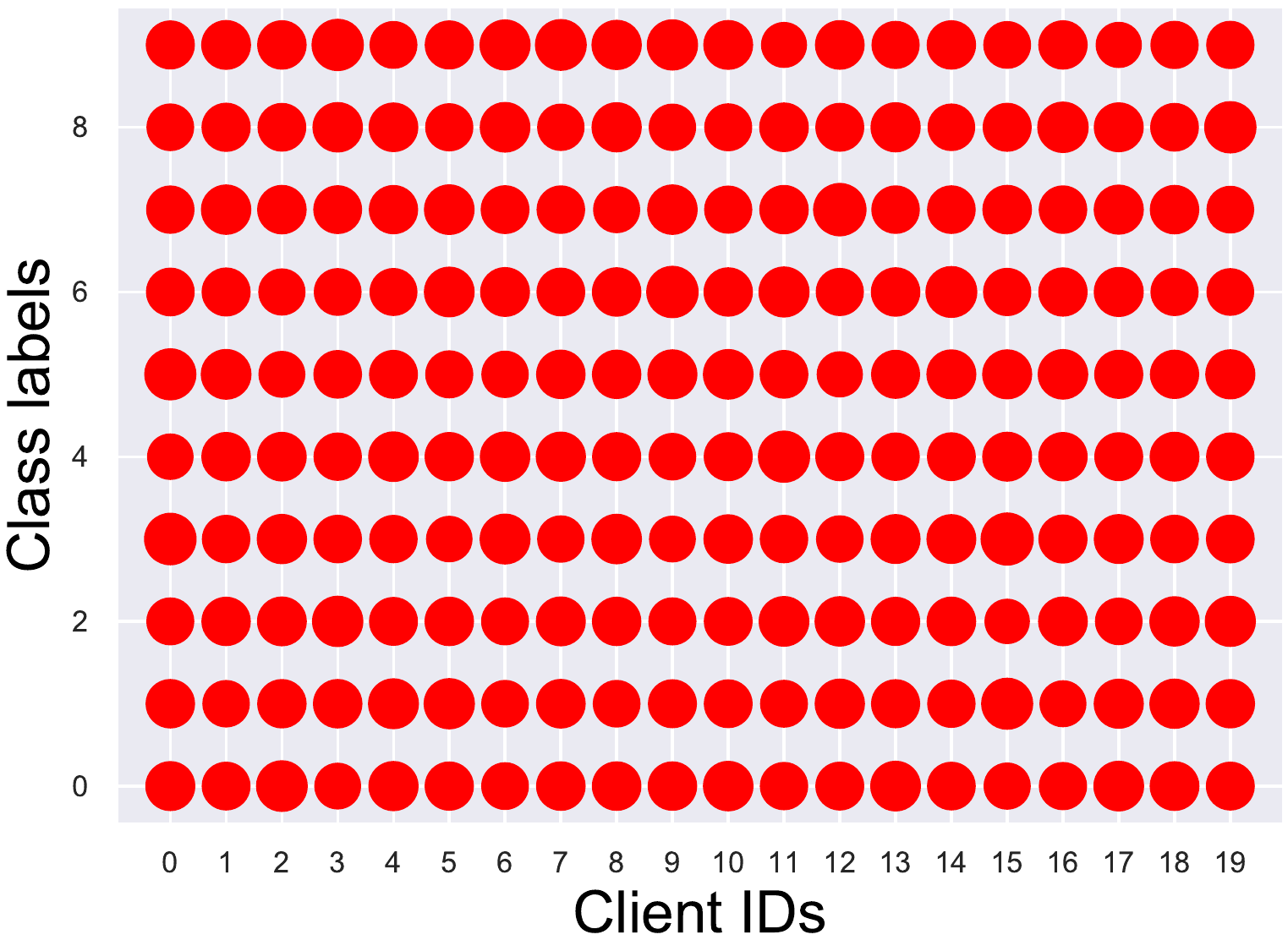}}
	\subfigure[CIFAR-10: $\alpha=1$]{\includegraphics[width=0.3\linewidth]{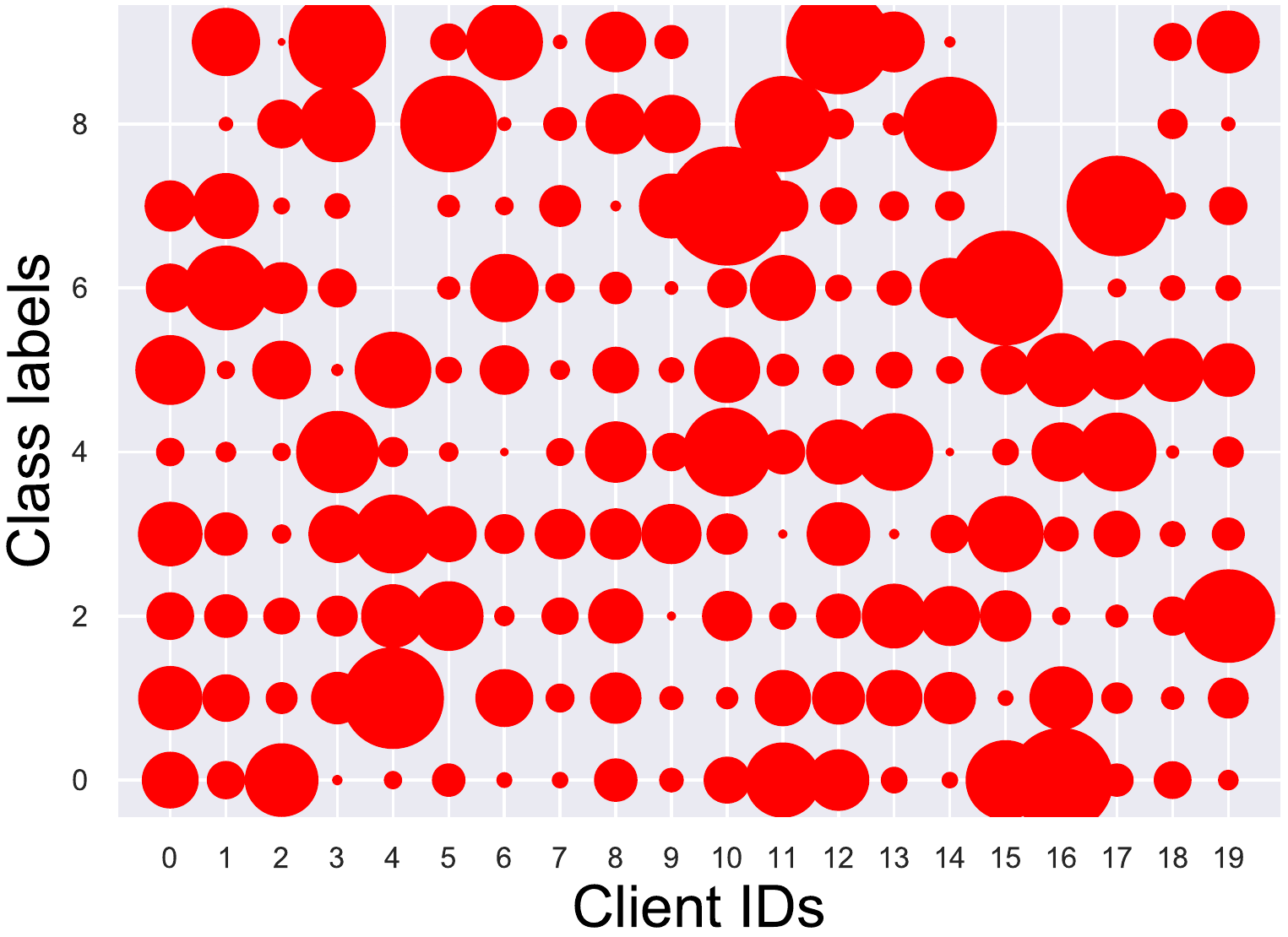}}
	\subfigure[CIFAR-10: $\alpha=0.5$]{\includegraphics[width=0.3\linewidth]{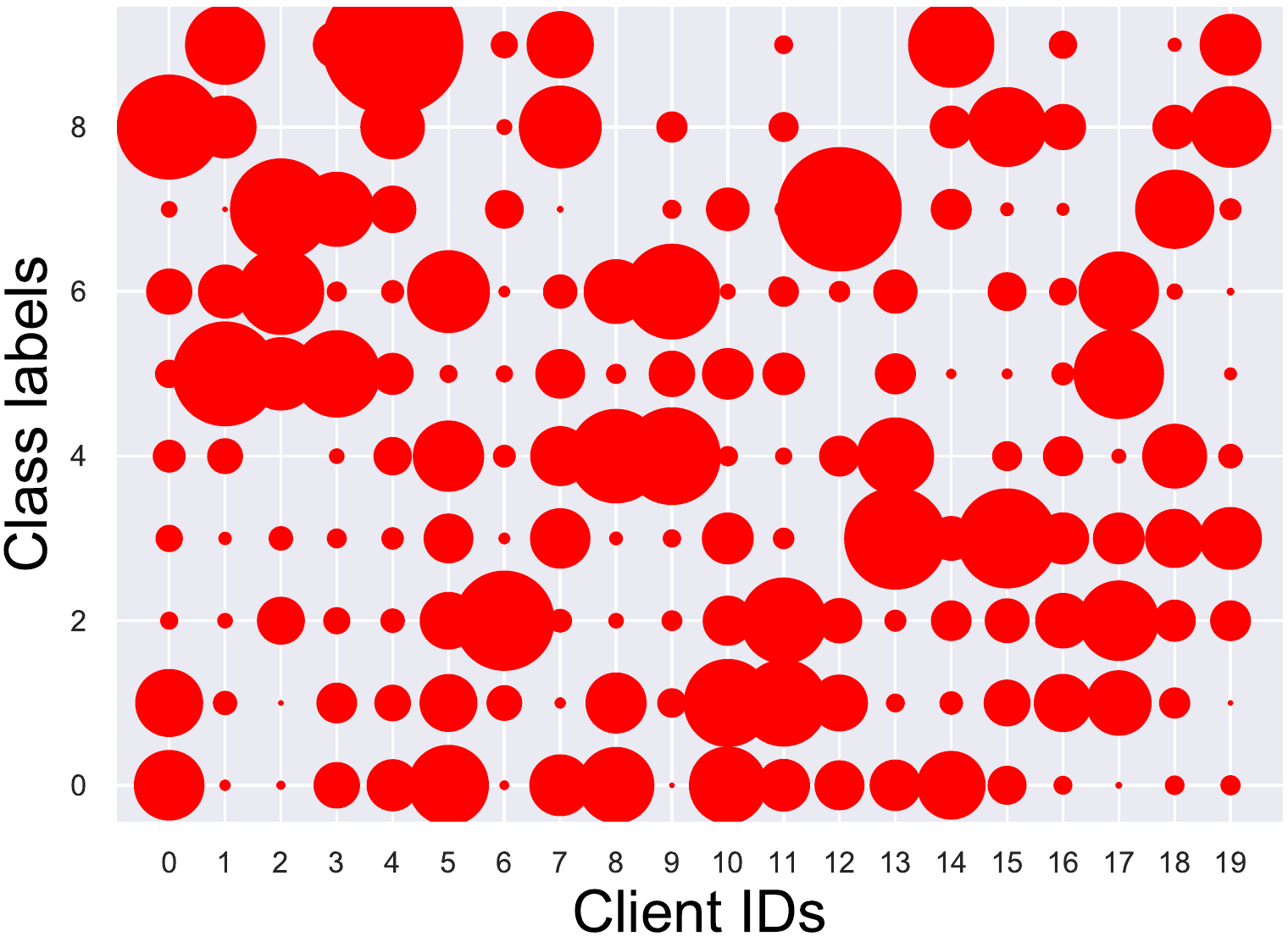}\label{cifar_05}}
	
	\subfigure[CINIC-10: $\alpha=1$]{\includegraphics[width=0.24\linewidth]{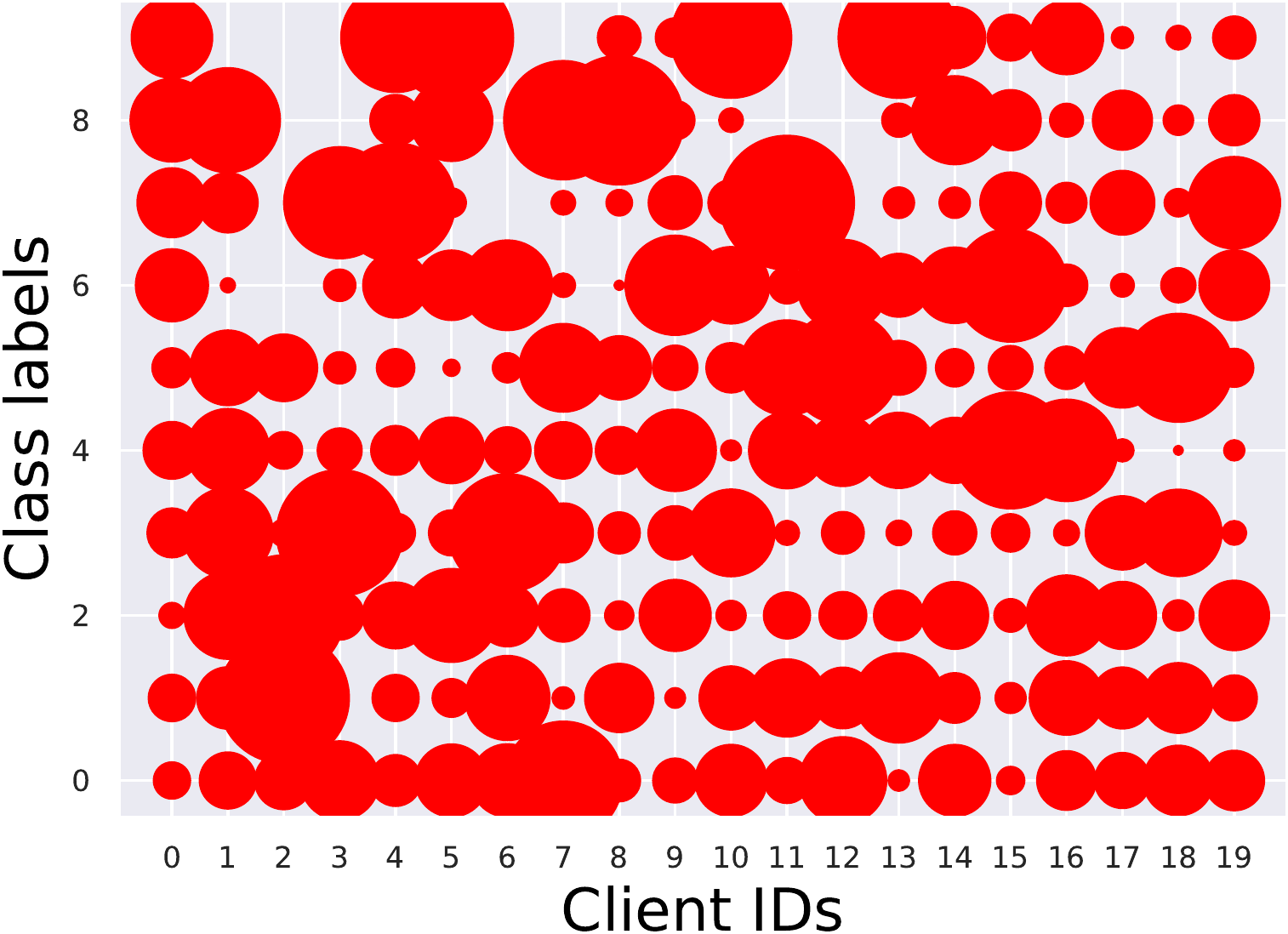}}
	\subfigure[CINIC-10: $\alpha=0.5$]{\includegraphics[width=0.24\linewidth]{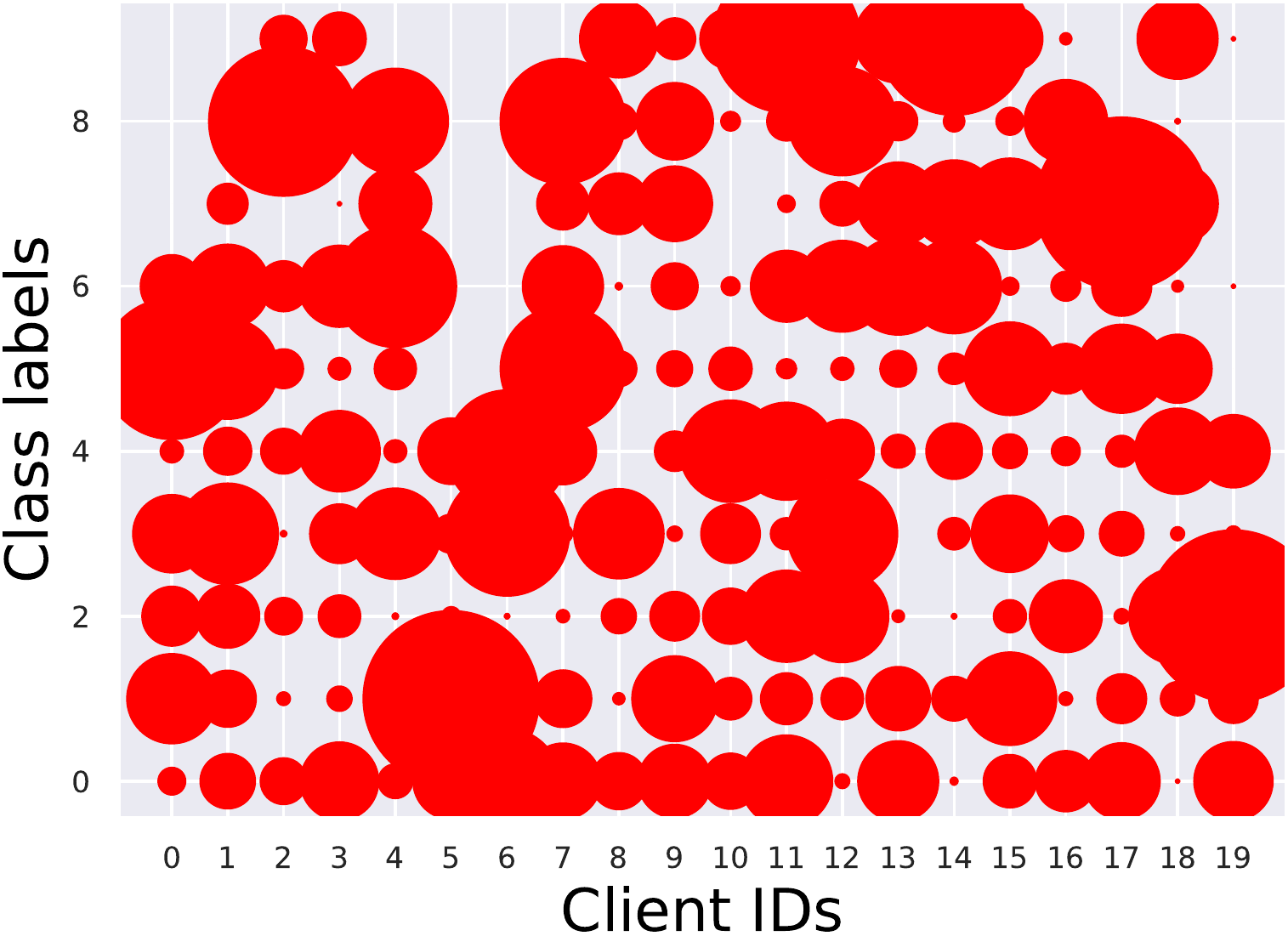}}
	\subfigure[CIFAR-100: $\alpha=1$]{\includegraphics[width=0.24\linewidth]{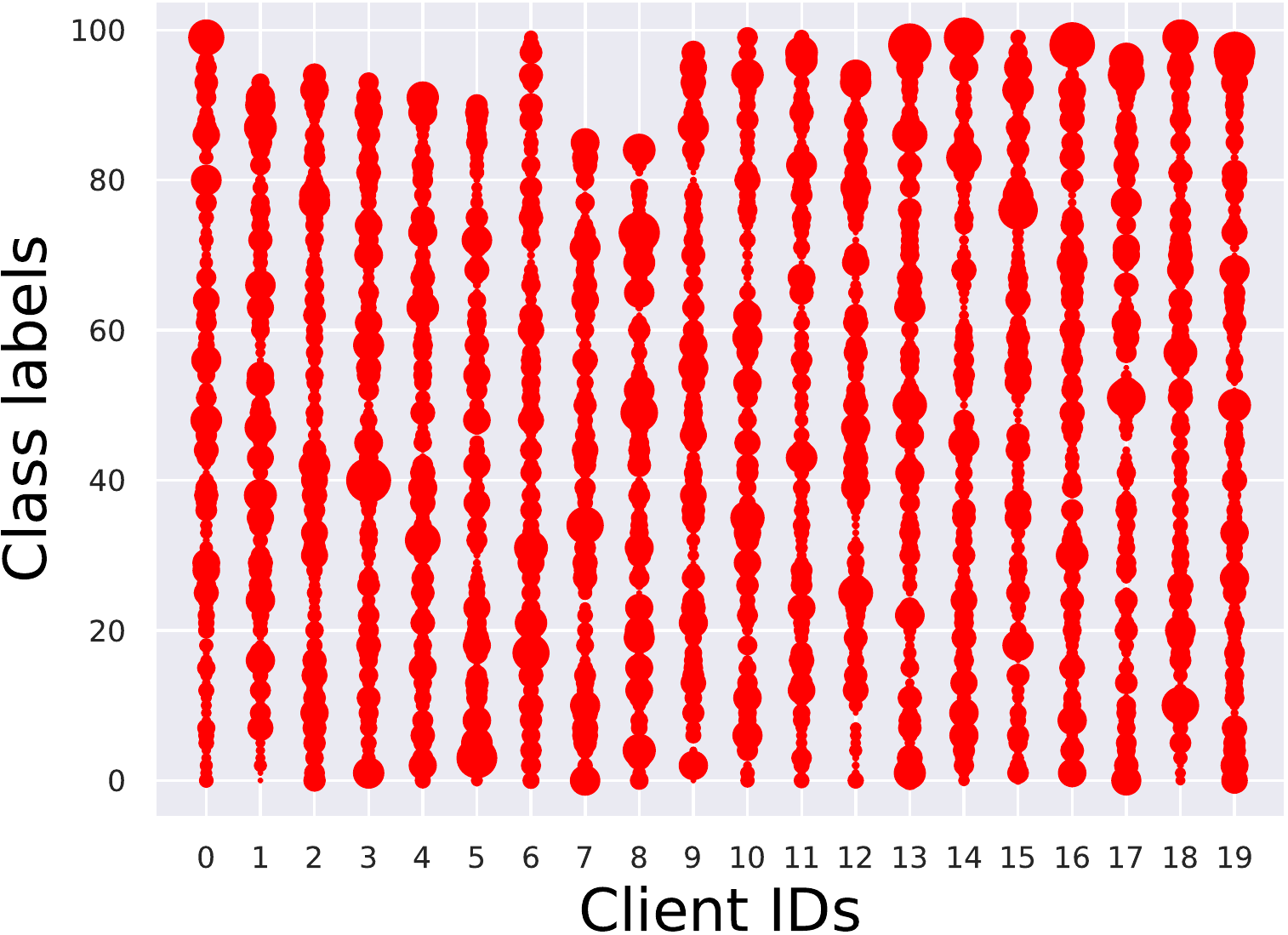}}
	\subfigure[CIFAR-100: $\alpha=0.5$]{\includegraphics[width=0.24\linewidth]{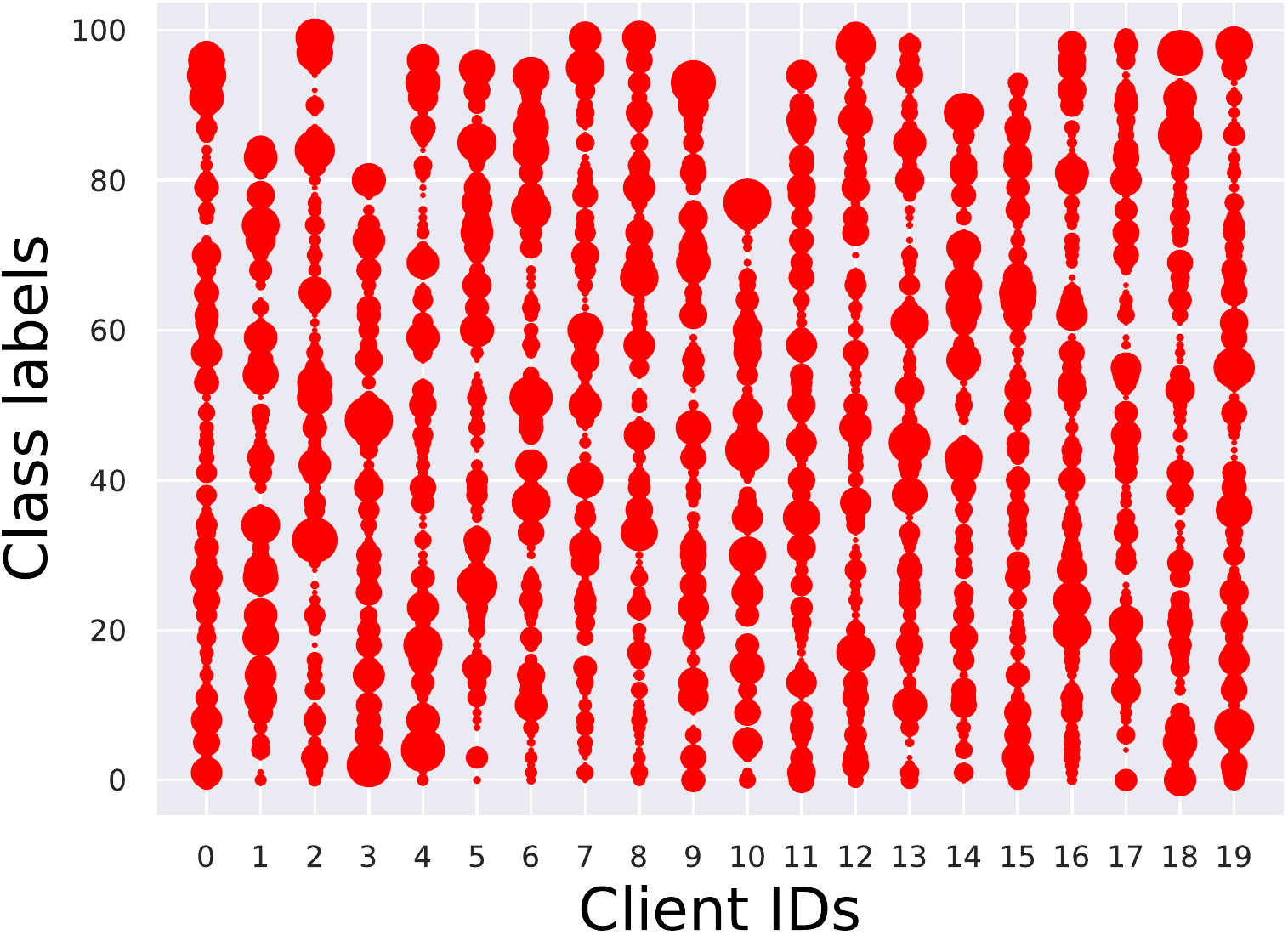}}
	\caption{{\bf Illustration of various degrees of non-IID data across clients} is sampled using Dirichlet distribution with $\alpha=\{100,1,0.5\}$ and indicated by dot sizes. The horizontal axis represents the total clients, and the vertical axis represents the classification labels.}
	\label{fig:non-iid}
\end{figure*}	

\section{Experiments} \label{exp}
In this section, we validate the performance of \texttt{RepPer} with non-IID data across clients from three aspects: 
(i) the discriminative of feature representations of the server and local clients learned in the CRL stage; 
(ii) the personalized classification performance for each client and generalization to the new client;
(iii) the effect of adapting to lower computation powers in representation and classification learning.

\subsection{Experiments Setup}
\subsubsection{Datasets with the heterogeneous distribution}
We consider the federated image classification problem using the following three real-world image datasets: CIFAR-10~\cite{cifar}, CIFAR-100~\cite{cifar}, and CINIC-10~\cite{cinic}. 
CIFAR-10/100 consist of 10/100 categories with 6000 and 600 samples per category, respectively. 
CINIC-10 is comprised of images from 10 categories with 9,000 samples per category.
We distribute these complete datasets to control heterogeneity using Dirichlet distribution as~\cite{FedDF, dirichlet:1, dirichlet:2}.
The concentration parameter $\alpha=\{100,1,0.5\}$ in Dirichlet distribution defines the degree of non-identicalness for client-partitioned data distribution. 
The possibility of the client holding samples from classes is positively correlated with the value of the concentration parameter: $\alpha= 100$ equals to identical distribution across all the clients, and as the $\alpha$ gets smaller, the clients are more likely to have samples from extremely class imbalance.
We consider three data partition strategies to simulate FL scenarios and visualize how samples distributed among 20 clients for CIFAR-10, CINIC-10, and CIFAR-100 on different $\alpha$ in Fig.~\ref{fig:non-iid}.

\subsubsection{Baselines}
We evaluate and compare against personalized FL methods as well as global shared methods with personalized fine-tuning. 
(1) \textbf{FedAvg}~\cite{fedavg} is the classical framework of FL and is treated as the baseline in this experiment. 
(2) \textbf{FedProx}~\cite{FRT:FedProx} leverages a regularization term to restrict local updates not far from the global model. 
(3) \textbf{LG-FedAvg}~\cite{Rep:LG-Fedavg} applies the representation learning strategy to learn local representation models and aggregates them into a global model to adapt to local data. 
(4) \textbf{FedRep}~\cite{Rep:FedRep} learns the global representation based on supervised cross-entropy loss and then updates personalized classifiers locally. 
(5) \textbf{FedAVG+FT} and \textbf{FedProx+FT}. For global shared models like FedAvg and FedProx, we train its global model first and then fine-tuning the classifier heads for 10 SGD epochs on its local training data, named \textbf{FedAVG+FT} and \textbf{FedProx+FT} for personalization and compute the final test accuracy. 

\subsubsection{Implementation}
In each experiment, all the baseline methods and \texttt{RepPer} share the same backbone network (ResNet34~\cite{resnet}), 
as well as the same number of epochs ($E=\{10, 20\}$ ), 
participant rate ($C=\{0.2,0.4,0.8\}$), 
batch sizes ($B=256$), 
learning rate ($\eta$=0.001, with a learning rate decay of 0.1.)
and communication round ($T=100$).
We use Adam~\cite{adam} as an optimizer, and the weight decay is set to $1\times{10}^{-4}$.
The value of hyperparameter temperature ($\tau$) in Eq.~\eqref{loss:CLR2} is fixed at 0.1.
Each global shared model baseline does for fine-turning its classifier head by running 10 local epochs of SGD. 
Personalized FL methods and PCL stage of \texttt{RepPer} did likewise. 
Federated classification accuracies are shown by averaging the local accuracy of each client on the corresponding test dataset. 
To better evaluate adaptation in realistic FL scenarios with heterogeneous settings, we show the accuracy of each client on non-IID test datasets. 
We implement our experiments based on the FedML~\cite{fedml}, an open-source federated learning library. 

\begin{figure*}[h]
	\centering
	\subfigure[Client A]{\includegraphics[width=0.2\textwidth]{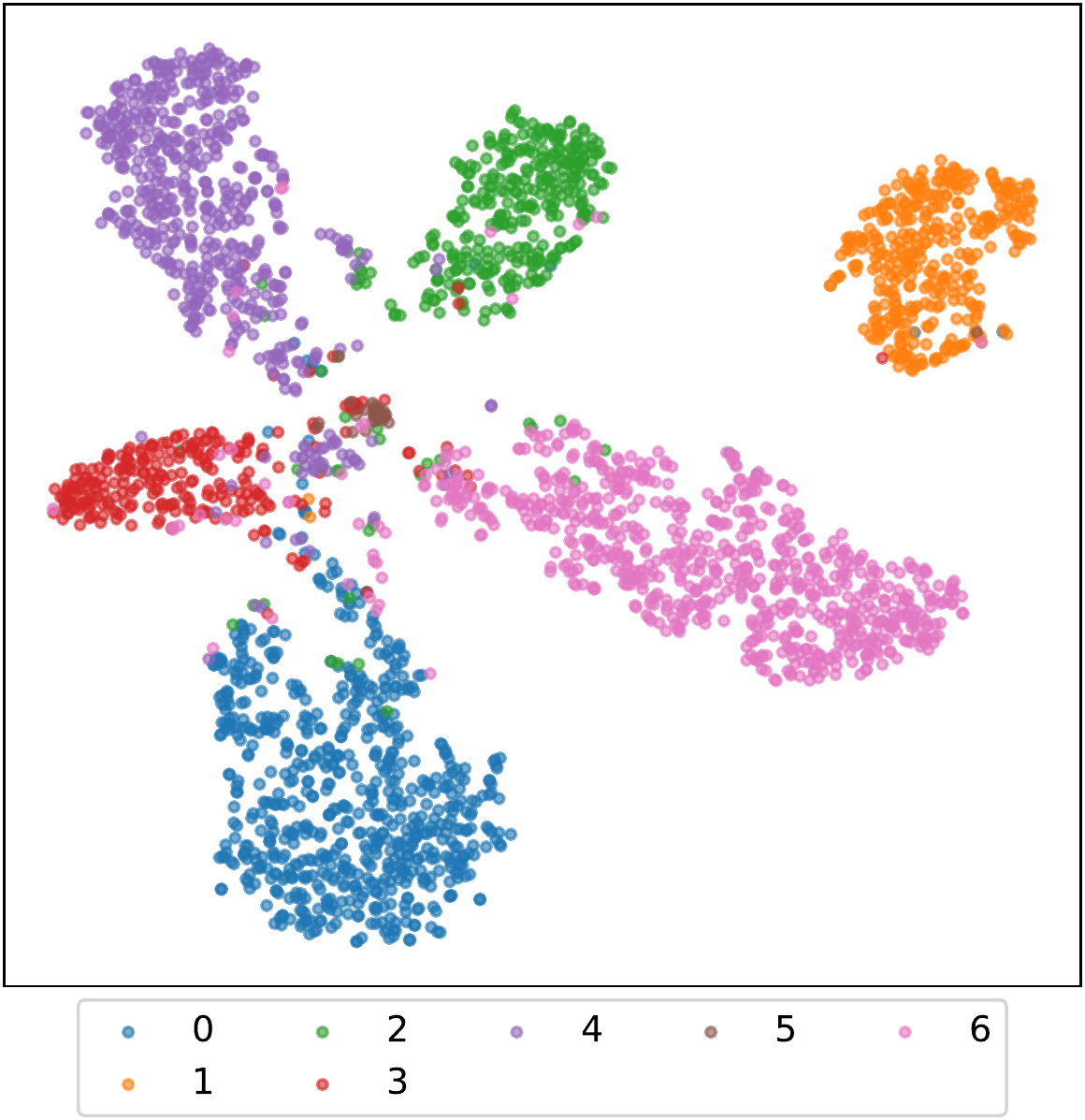}\label{tsne:a}}
	\subfigure[Client B]{\includegraphics[width=0.2\textwidth]{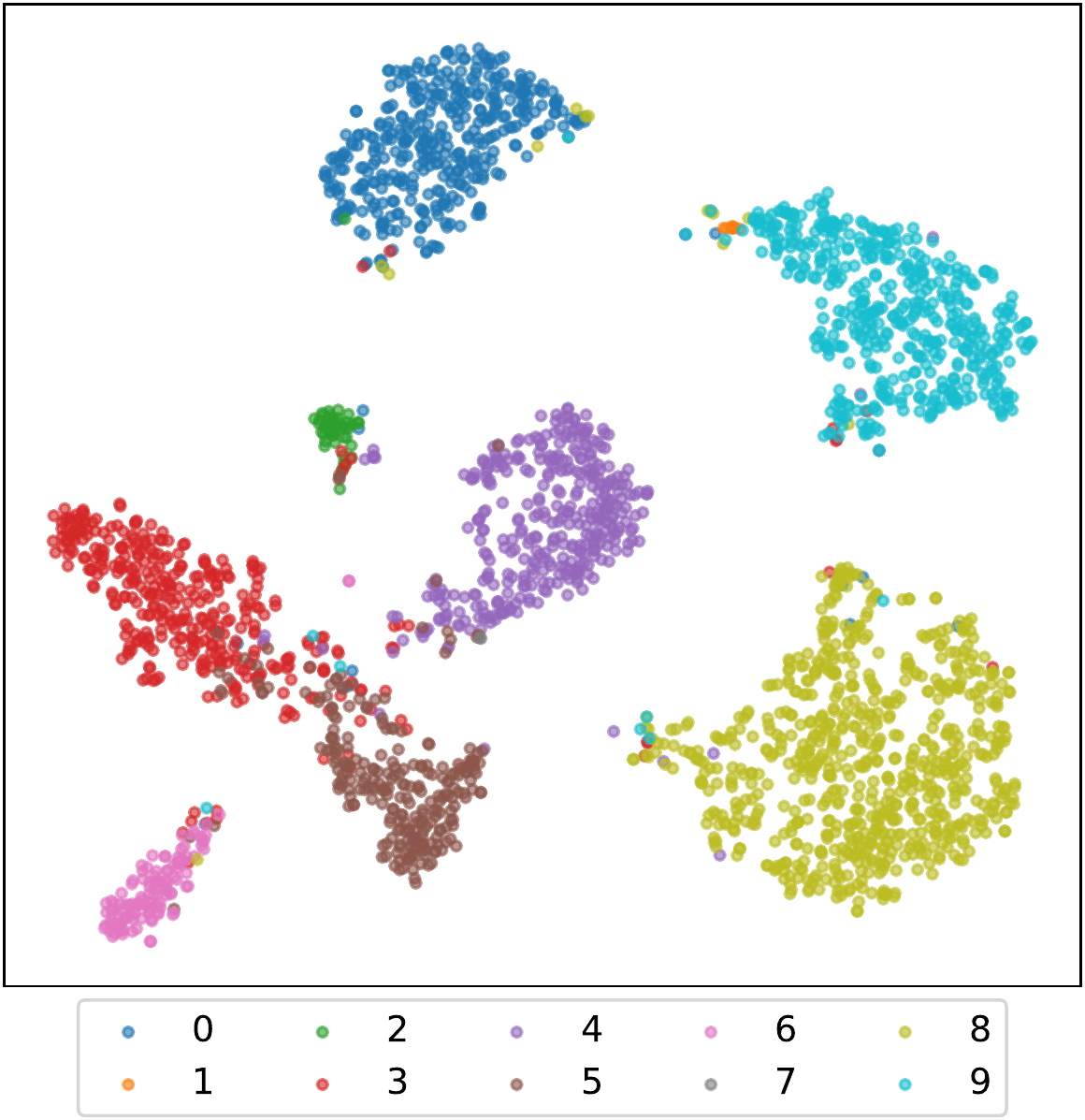}\label{tsne:b}}
	\subfigure[Client C]{\includegraphics[width=0.2\textwidth]{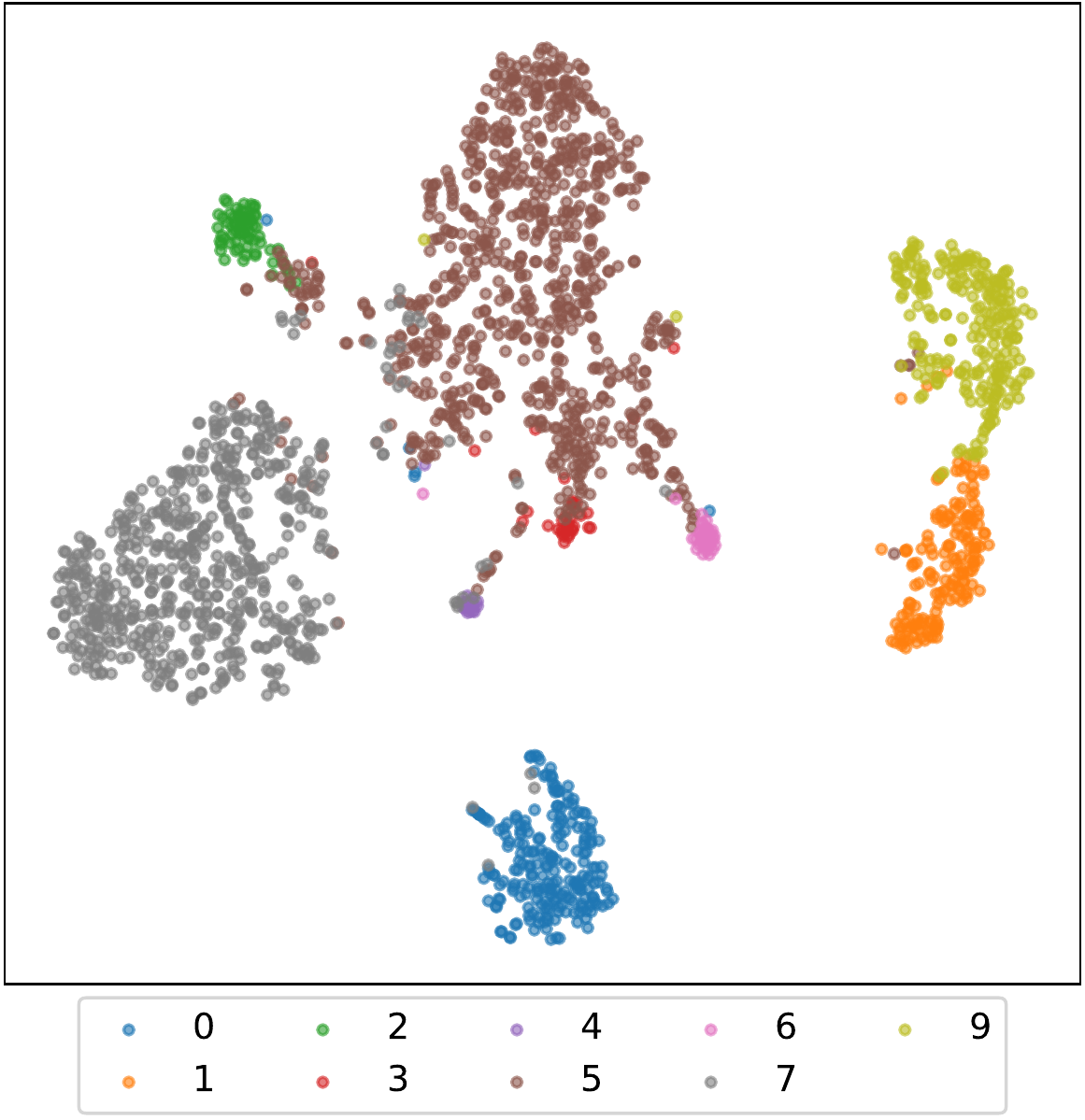}\label{tsne:c}}	
	\subfigure[Client D]{\includegraphics[width=0.2\textwidth]{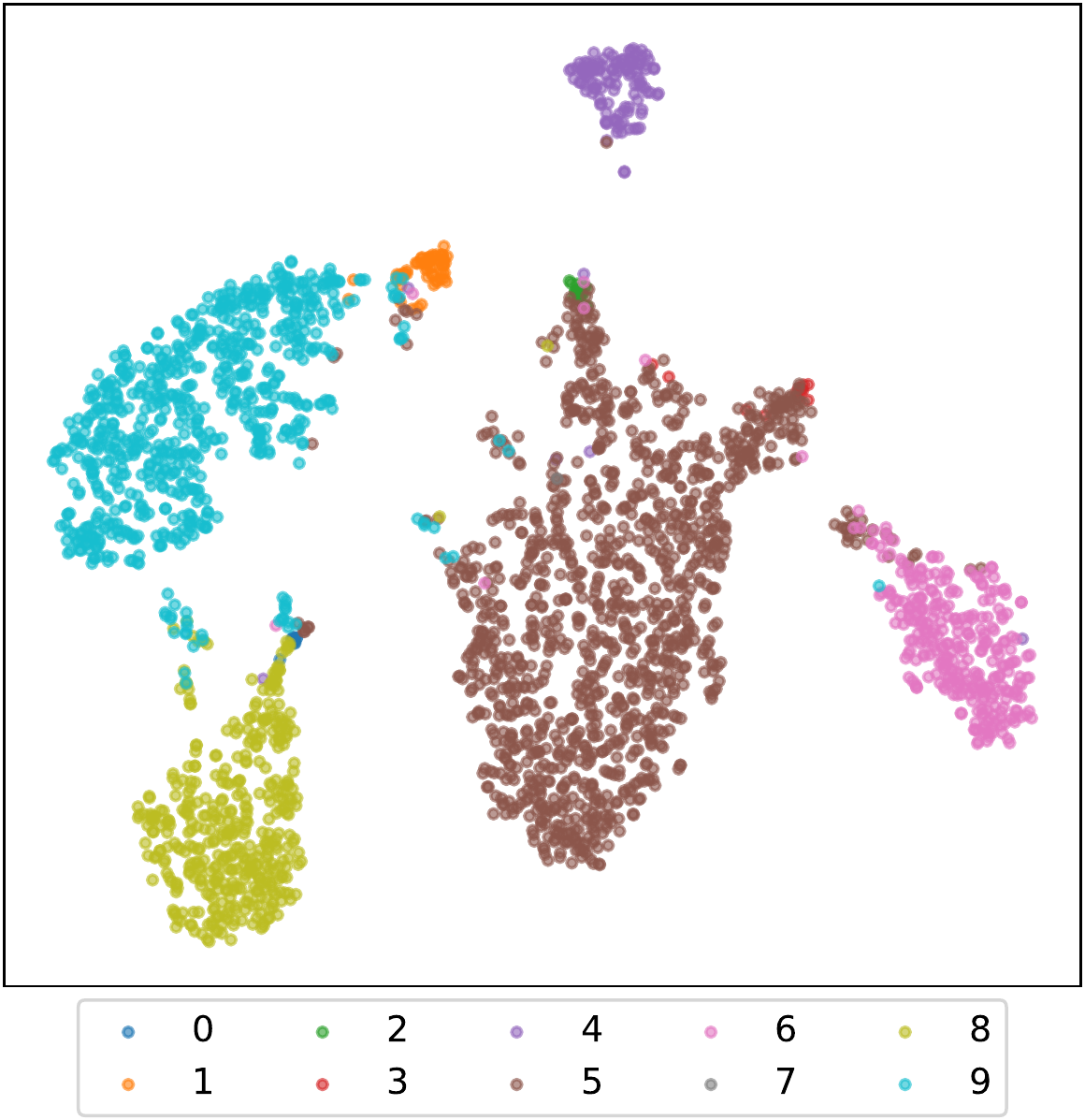}\label{tsne:d}}
	
	\subfigure[Client E]{\includegraphics[width=0.2\textwidth]{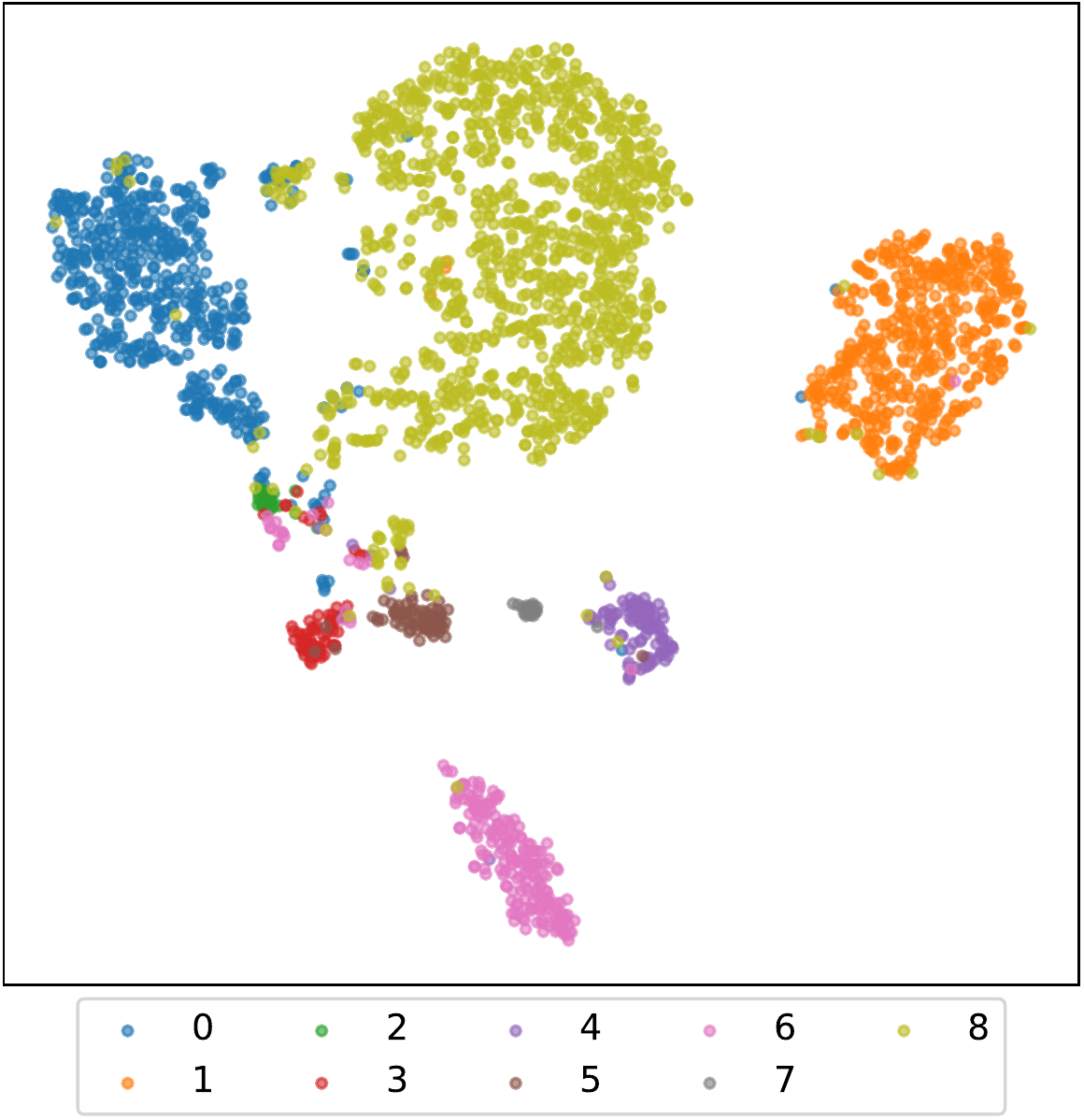}\label{tsne:e}}
	\subfigure[Client F]{\includegraphics[width=0.2\textwidth]{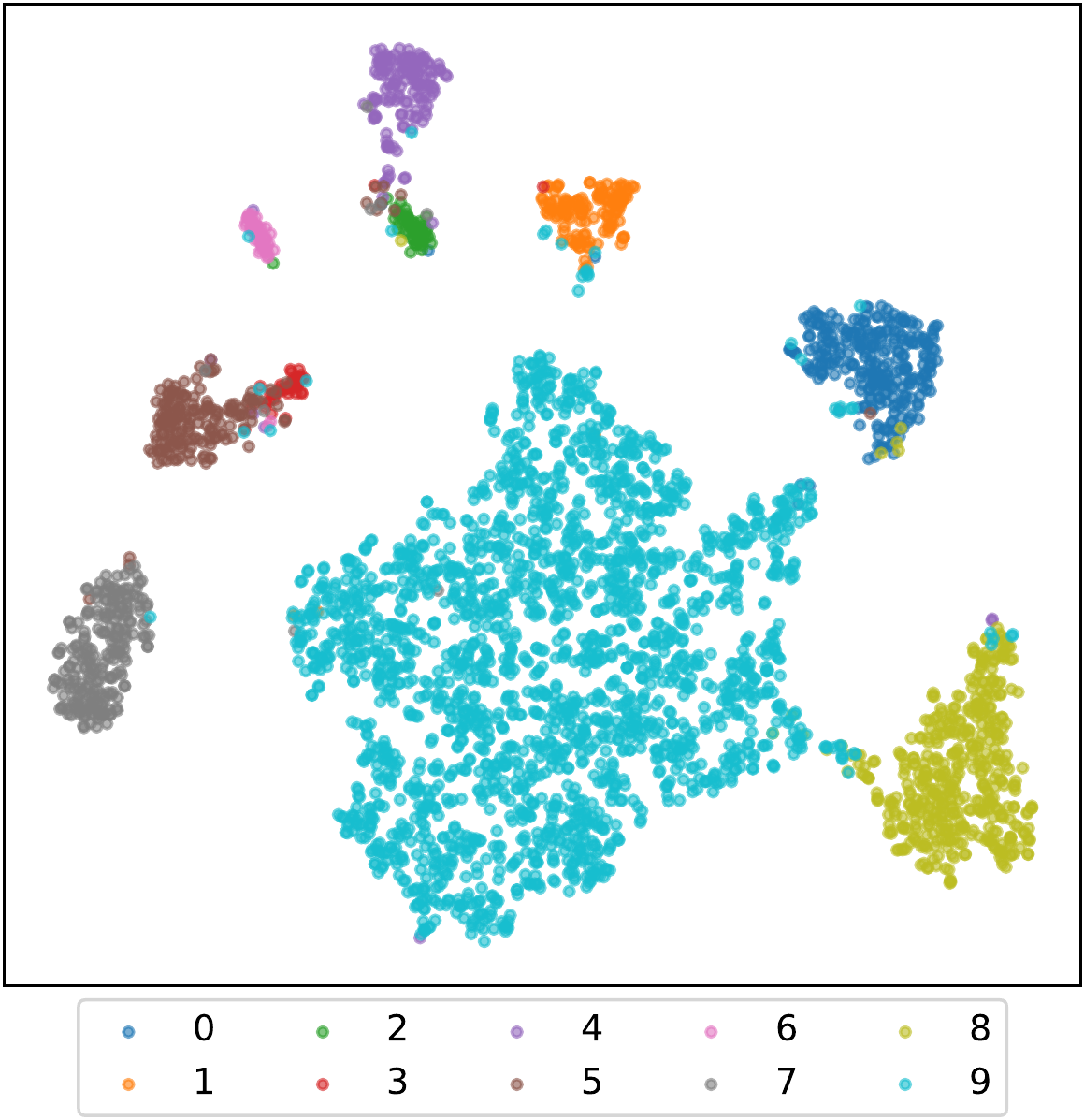}\label{tsne:f}}
	\subfigure[Client G]{\includegraphics[width=0.2\textwidth]{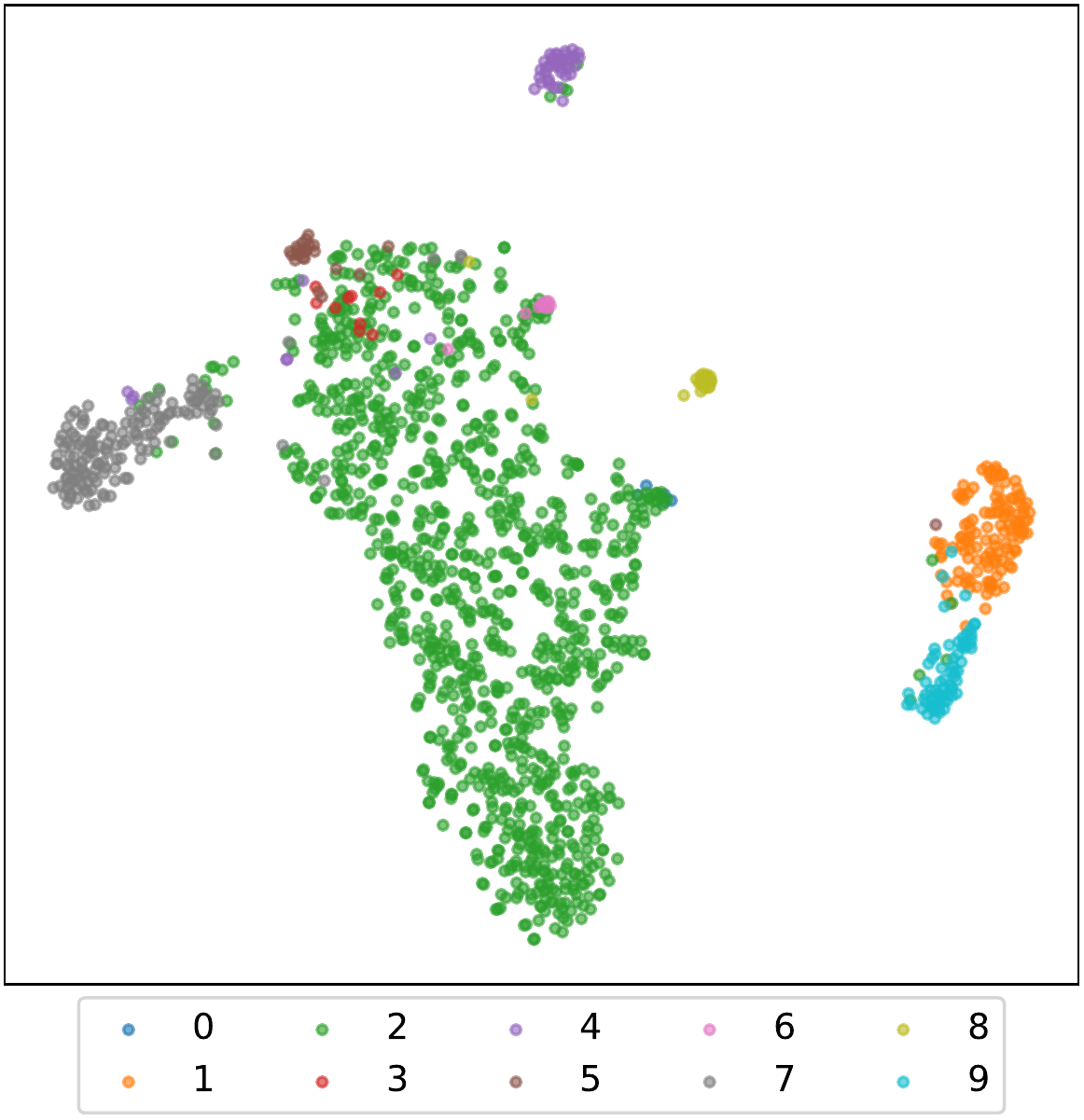}\label{tsne:g}}
	\subfigure[Server]{\includegraphics[width=0.2\textwidth]{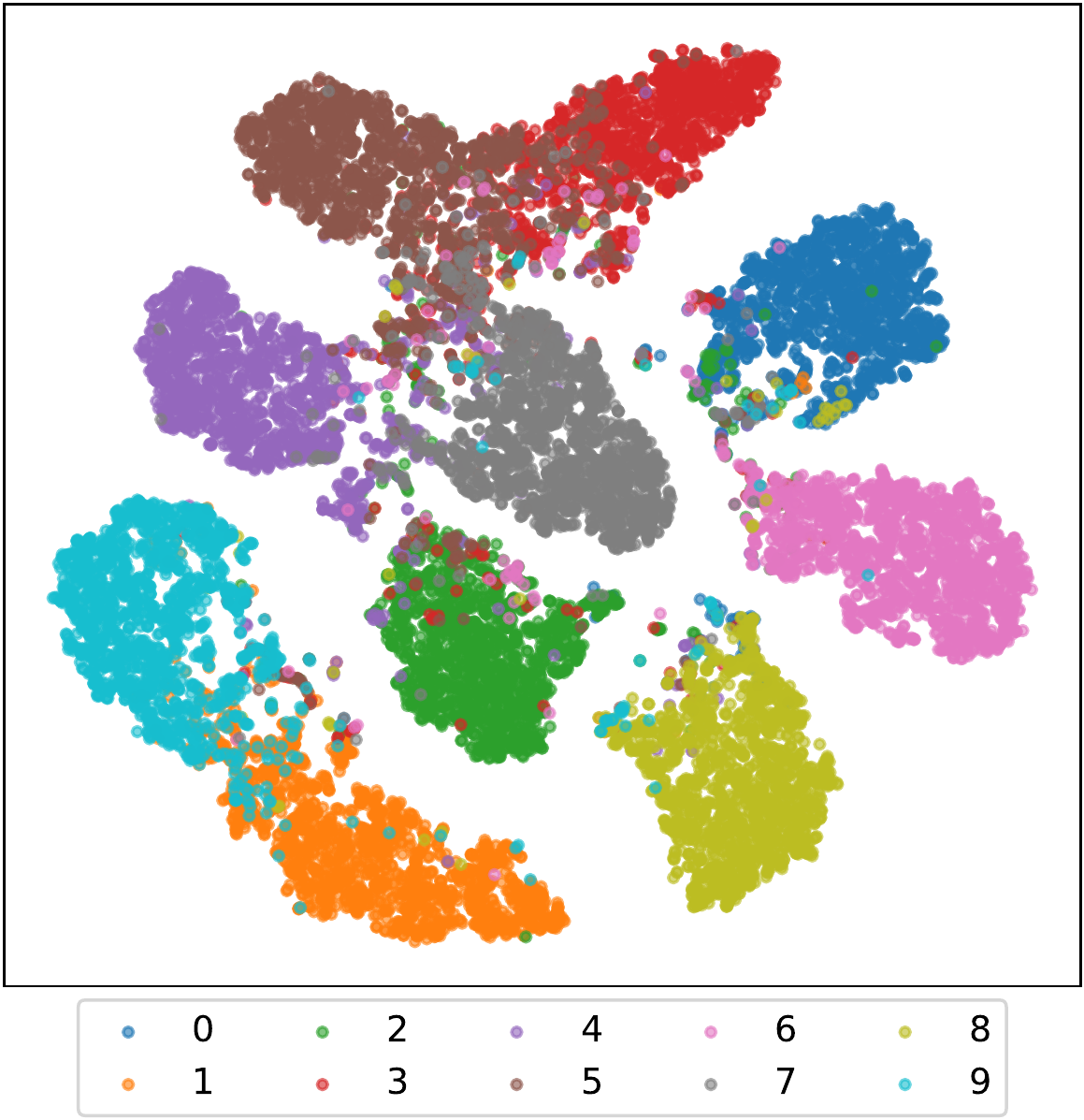}\label{tsne:s}}
	\caption{{\bf T-SNE visualization of representations learned from global and local models in \texttt{RepPer} on the CIFAR-10 under non-IID setting with $\alpha=0.5$.} 
		For local data, each client contains an unequal number or category of images, and the number of categories in the enumerated randomly selected clients is shown in the respective legends.
		From left to right, the data distribution becomes increasingly skewed. 
		Client A includes samples from 7 categories, and the number of samples per category is relatively equal.
		Clients B, C, and D contain a small number of tail categories and a relatively large number of samples from the remaining categories.
		Clients E, F, and G include samples with extremely class imbalance, containing some categories with minimal numbers.
		The server aggregates and updates with local models, achieving well-clustering results on the whole data.
	}\label{fig:tsne}
\end{figure*}

\subsection{Representations Visualization}\label{sec:rep}
To validate the feature representation capability of the CRL stage in \texttt{RepPer}, we visualize representations learned from global and local local models using t-Distributed Stochastic Neighbor Embedding (t-SNE)~\cite{tsne}, as shown in Fig.~\ref{fig:tsne}.
Different colors in the figure represent different data categories. 

We first show representations of the seven randomly selected clients on their respective local data. 
In the non-IID setting of $\alpha=0.5$,  as shown in Fig.~\ref{cifar_05}, local data exhibit varying class imbalances among the clients.
Despite this, we observe that local models can cluster well for various degrees of instance-rich/scarce classes, leading to sharper boundaries and better discrimination of tail categories in t-SNE.
In \texttt{RepPer}, the server collects local representation model parameters to update a global representation model. 
We then further show representation on the server over all the data and we observe that the global representation model consistently performs well even though the local representation models are trained on the non-IID data.

Fig.~\ref{fig:tsne} illustrates that \texttt{RepPer} learns local representing models on heterogeneous data and obtains a common global model unaffected by non-IID data. 
We further show evaluation of the effectiveness of personalized classification based on the common representation model in the next section.

\subsection{Model Performance}
In this section, we performed a detailed comparison of \texttt{RepPer} and baselines on federated image classification. 
We first investigate the effect of hyperparameters on the classification accuracy of FedAvg, LG-FedAvg, FedRep, and \texttt{RepPer}, in both IID and various degrees of the non-IID settings. 
Then, we compare the performance of \texttt{RepPer} with baselines in the non-IID setting. 
Finally, the out-of-local-distribution generalization of \texttt{RepPer} is compared to that of FL methods.

\subsubsection{Effect of hyperparameters}
The key factors considered in FL that affect performance include: the fraction ($C$) of clients participating on each round, the number of local epochs  ($E$), and degrees of heterogeneity ($\alpha$) in the dataset. 
To understand the critical factors that affect the convergence of \texttt{RepPer} in both IID and non-IID settings, we evaluate \texttt{RepPer} and compare it with baseline methods on the CIFAR-10 dataset.
We compare the test performance of the proposed \texttt{RepPer} with baseline methods for different local training settings: non-IID degrees $\alpha= \{100,1,0.5\}$; 
randomly select a fraction $C= \{0.2,0.4,0.8\}$ of the total $20$ clients;
the number of communications is $100$, and local epochs $E= \{10,20\}$ per communication round. 
The various values of the three critical parameters and their effects on the methods are described in Table~\ref{fig:hyper}.

\begin{itemize}
	\item 
	\textbf{Effect of ${\alpha}$.} 
	Statistic heterogeneous is the main factor affecting the performance of FL models. 
	In this experiment, only the value of $\alpha$ reduces from $100$ to $1$ and finally to $0.5$, we notice that almost all the baseline methods are suffered from statistic heterogeneous on classification.
	This is because the model aggregation is highly dependent on local models. 
	The non-IID data lead to biased local models and inferior global model. 
	However, \texttt{RepPer} still allows high accuracy of the continuous classification.
	The reason is that \texttt{RepPer} focuses on learning representation clusters instead of decision boundaries for classification on non-IID data, which is impervious to label bias and client-drift, as illustrated in Fig.~\ref{fig:tsne}.
	
	\item 
	\textbf{Effect of ${C}$.} 
	The fraction ($C$) of clients controls parallelism in each communication round. 
	We show the effect of varying $C$ for baseline models and \texttt{RepPer}. 
	According to Table~\ref{fig:hyper}, the down arrow ($\downarrow$) represents a decrease in classification accuracy as the number of clients participating increases.
	We observe that increasing $C$ may not always be positively correlated with performance improvement of baseline methods, but has a limited impact on \texttt{RepPer}, neither in the IID nor non-IID cases. 
	This observation reminds us that not all methods benefit from increasing the number of participating clients, and the value of $C$ should be carefully chosen for the remaining experiments.
	
	\begin{table*}
		\caption{{\bf Evaluation of different FL approaches in various heterogeneous settings with three key parameters}: client sampling fractions $C$, local epochs $E$, and non-IID degrees $\alpha$. 
			We show classification test accuracy, and the best results are in bold. 
			We adopt the same backbone as ResNet-34 for each method and evaluate the performance on the CIFAR-10 dataset.}
		\label{fig:hyper}
		\centering
		\resizebox{\textwidth}{!}{
			\scalebox{0.70}{
				\begin{tabular}{*{11}{c}}
					\toprule[1pt]
					\multirow{2} * {Method} & \multirow{2} * {Local epochs} & \multicolumn{3}{c}{$C=0.2$}  & \multicolumn{3}{c}{$C=0.4$} & \multicolumn{3}{c}{$C=0.8$}\\
					\cmidrule(lr){3-5} \cmidrule(lr){6-8} \cmidrule(lr){9-11}
					& & IID & $\alpha =1.$ & $\alpha =0.5.$ & IID & $\alpha =1.$ & $\alpha =0.5.$ & IID & $\alpha =1.$ & $\alpha =0.5.$\\
					\midrule[1pt]
					\multirow{2}*{FedAvg~\cite{fedavg}} & 10 &78.04 &72.58 &66.56 &78.58 &72.64 &70.91 &79.05 &74.73 &75.64\\
					& 20 &77.61 &72.39 &66.43 &78.28 &73.65 &71.67 &78.66 &74.90 &75.99\\         
					\cmidrule(lr){1-11} 
					\multirow{2}*{LG-FedAvg~\cite{Rep:LG-Fedavg}} & 10 &78.87 &75.67 &73.62 &78.25$\downarrow$  &74.68 &74.88 &78.03$\downarrow$ &74.53$\downarrow$ &75.88\\
					& 20 &77.25 &74.27 &73.82 &76.81$\downarrow$ &73.90$\downarrow$ &72.98$\downarrow$ &77.03 &74.06 &74.26\\ 
					\cmidrule(lr){1-11} 
					\multirow{2}*{FedRep~\cite{Rep:FedRep}} & 10 &80.43 &67.04 &61.08 &80.53 &73.03 &76.36 &79.86 &74.57 &75.49$\downarrow$\\
					& 20 &78.21 &71.24 &71.83 &79.64 &69.45$\downarrow$ &71.35$\downarrow$ &80.03 &74.21 &71.85\\
					\cmidrule(lr){1-11} 
					\multirow{2}*{{\bf \texttt{RepPer}}} & 10  &{\bf 90.62} &{\bf 84.98} &{\bf 81.04} &{\bf 91.16} &{\bf 86.22} &{\bf 82.17} &{\bf 90.84}$\downarrow$ &{\bf 86.63} &{\bf 82.48}\\
					& 20 &{\bf 91.90} &{\bf 86.88} &{\bf 83.48} &{\bf 91.57} &{\bf 87.71} &{\bf 83.18} &{\bf 91.80} &{\bf 87.73} &{\bf 83.11}$\downarrow$\\
					\bottomrule[1pt]
				\end{tabular}
		}}
	\end{table*}
	
	\begin{table*}[ht]
		\renewcommand\arraystretch{1.25}
		\caption{{\bf Top-1 accuracy $(\%)$ comparison on non-IID settings of CINIC-10, CIFAR-10/100 datasets.} 
			We keep randomly selected the fraction $C=0.2$ of $20$ clients and epochs $E=10$ in iterations constant and evaluated the performance of SOTA baselines and \texttt{RepPer} on degrees of non-IID data, which is synthetically controlled by $\alpha=\{1,0.5\}$. 
			Best results are marked in \textbf{bold}, and suboptimal results are marked with underline.
			\texttt{RepPer} outperforms prior work across non-IID settings. }
		\label{table:comparison}
		\begin{center}
				\begin{tabular}{*{10}{lc}}
					\toprule[1pt]
					\multirow{2} * {Method} & \multicolumn{2}{c}{\bf CINIC-10}  & \multicolumn{2}{c}{\bf CIFAR-10} & \multicolumn{2}{c}{\bf CIFAR-100}\\
					\cmidrule(lr){2-3} \cmidrule(lr){4-5} \cmidrule(lr){6-7}
					& $\alpha =1.$ & $\alpha =0.5.$ & $\alpha =1.$ & $\alpha =0.5.$  & $\alpha =1.$ & $\alpha =0.5.$\\
					\midrule[1pt]
					FedAvg~\cite{fedavg}          &57.77 &56.73 &72.58 &66.56 &40.87 &40.21 \\
					FedAvg+FT          			  &60.90 &60.91 &72.62 &69.41 &41.35 &40.89 \\
					FedProx~\cite{FRT:FedProx}    &\underline{65.09} &63.28 &72.75 &72.88 &43.00 &41.53\\
					FedProx+FT                           &64.31 &\underline{63.69} &72.85 &72.92 &43.19 &41.79\\
					FedRep~\cite{Rep:FedRep}         &59.45 &59.33 &67.04 &61.08 &40.06 &36.36 \\
					LG-FedAvg~\cite{Rep:LG-Fedavg}       &59.95 &59.12 &\underline{75.67} &\underline{73.62} &\underline{45.08} &\underline{44.85} \\
					{\bf \texttt{RepPer}}  &{\bf 74.67} &{\bf 65.09} &{\bf 84.98} &{\bf 81.04} &{\bf 55.38} &{\bf 53.86} \\
					\bottomrule[1pt]
				\end{tabular}
			\end{center}
		\end{table*}	
	
	\item 
	\textbf{Effect of ${E}$.} 
	Intuitively, small $E$ increases the communication burden. 
	Meanwhile, more local epochs ($E$)	typically lead to divergence in FL. 
	In this experiment, we fix $C$ and $\alpha$, and increasing $E$ shows that the larger value of $E$ brings negative effects on almost all the baseline methods, which is consistent with observations in~\cite{fedavg, RN20, leaf}. 
	The longer training epochs bring greater client diversity, resulting in inferior global model aggregation. 
	In contrast, \texttt{RepPer} can benefit from more local epochs ($E$).
	\texttt{RepPer} focuses on learning representations from non-IID local data, and more local training epochs boost representation ability to target various heterogeneous FL scenarios. 
	Therefore, as the local epochs increases, the performance of \texttt{RepPer} is improved. 
\end{itemize}
The analysis of these key factors shows that \texttt{RepPer} can benefit from more local epochs ($E$), a specific number of participant clients ($C$), and robustness on various non-IID data with $\alpha$, and result in a steady rise in federated classification accuracy.

	\subsubsection{Performance Comparison}
	In order to highlight the personalization performance of the \texttt{RepPer}, we continue to compare it with more FL methods, including FedAvg, FedProx, LG-FedAvg, FedRep, and the global model with fine-tuning the classifier heads for personalization, including FedAvg+FT and FedProx+FT.
	In all scenarios, we consider the heterogeneous setting including: degrees of non-IID with $\alpha=\{1,0.5\}$, a fraction $C= 0.2$ of the total 20 clients are randomly selected in iteration, 100 communication rounds with 10 local epochs in each. 
	
	Table~\ref{table:comparison} shows the top-1 classification accuracies (with the same hyperparameters) on the various non-IID degrees of CINIC-10, CIFAR-10, and CIFAR-100 datasets. 
	The comparison on the CINIC-10 dataset shows that \texttt{RepPer} in non-IID degrees $\alpha=\{1,0.5\}$ have $9.58\%$ and $1.4\%$ more accurate than the best performing alternatives, FedProx and FedProx+FT, respectively.
	For the CIFAR-10 and CIFAR-100 datasets, we observe the best-performing alternative on non-IID degrees $\alpha=\{1,0.5\}$ is LG-FedAvg. 
	Table~\ref{table:comparison} shows that \texttt{RepPer} has a $9.31\%$ and $7.42\%$ improvement in accuracy than LG-FedAvg on the CIFAR-10 dataset.
	\texttt{RepPer} achieves $10.3\%$ and $9.01\%$ more accuracy on CIFAR-100 datasets than LG-FedAvg.  
	
\begin{figure*}[ht]
		\centering
		\subfigure{\includegraphics[width=0.5\columnwidth]{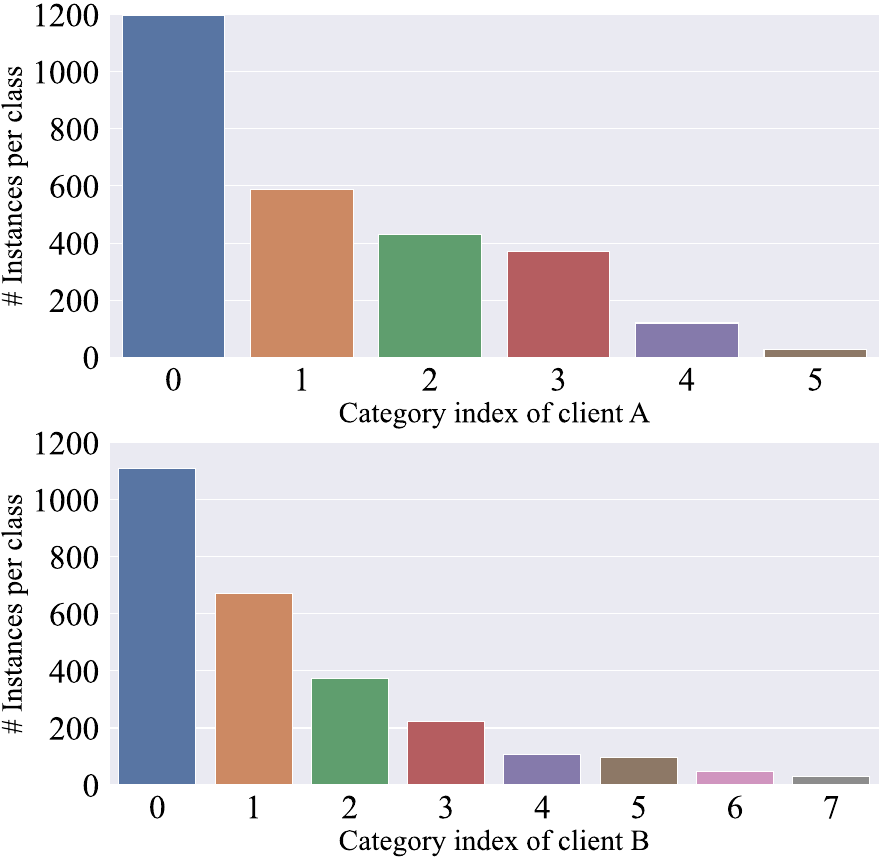}}
		\subfigure{\includegraphics[width=0.7\columnwidth]{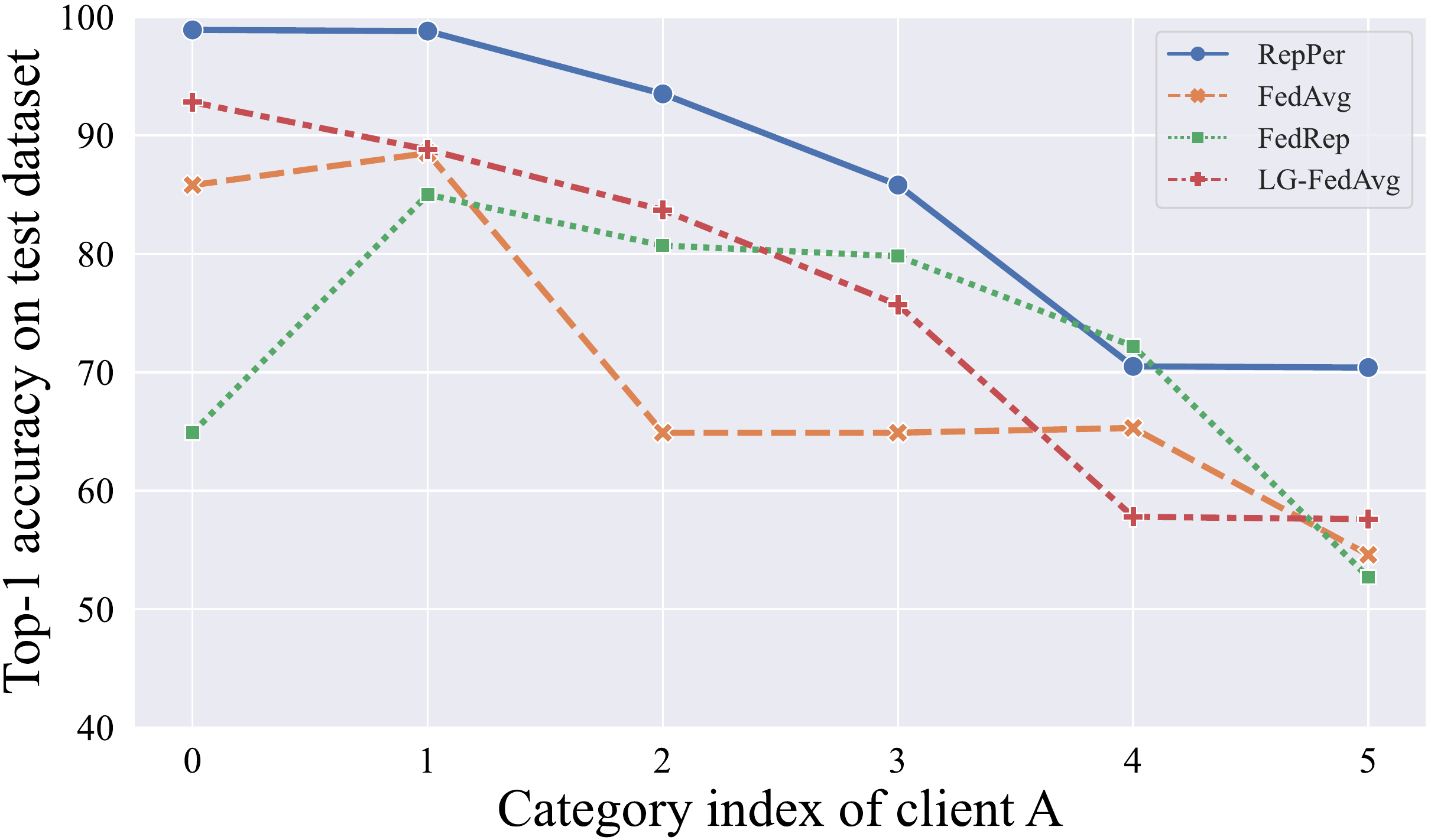}}
		\subfigure{\includegraphics[width=0.7\columnwidth]{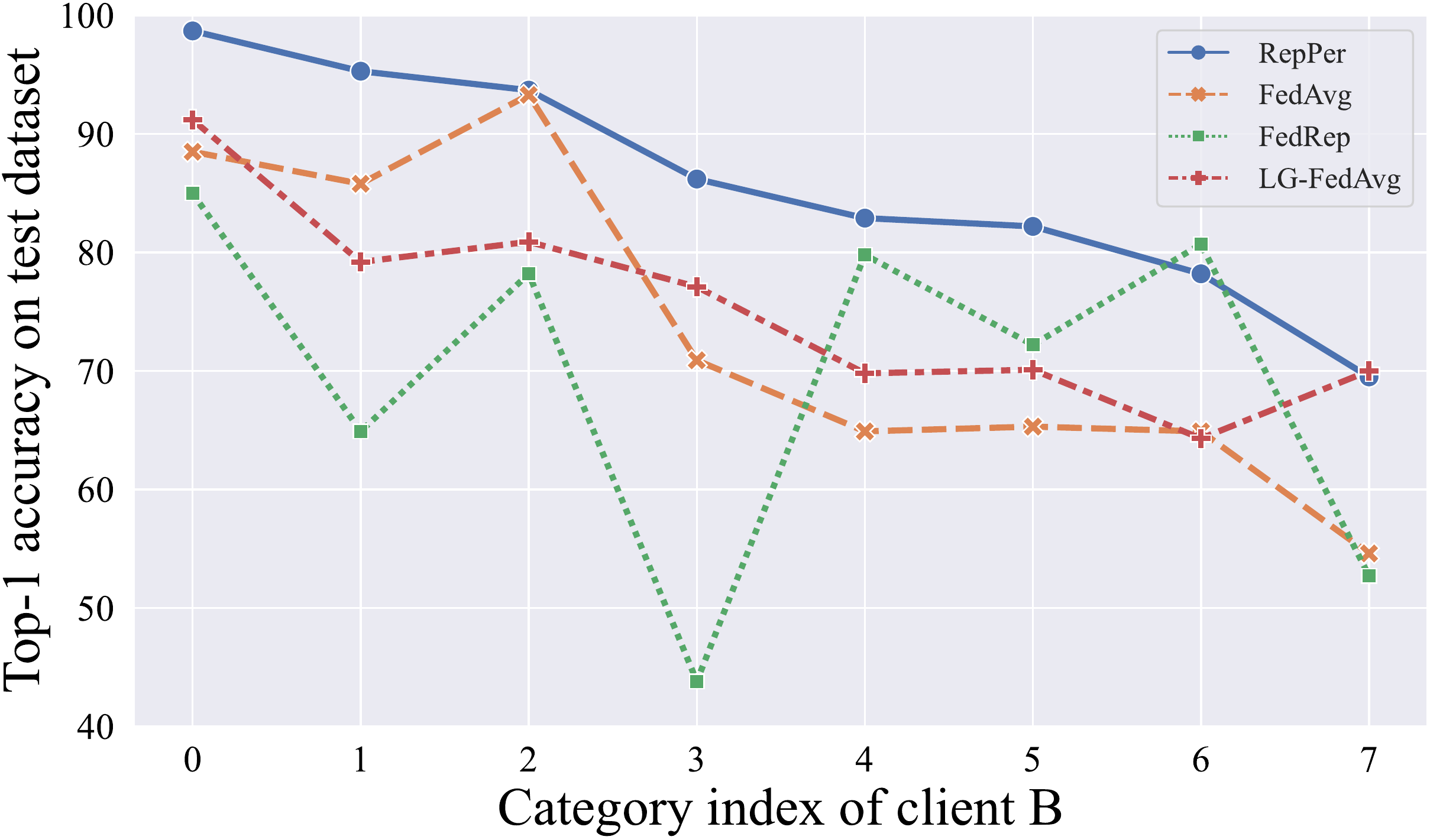}}
		\caption{\textbf{Classification accuracy of two randomly selected clients in their respective classes. }
			Each client contains local training data with different categories, and category indexes are sorted by the number of samples in each category in descending order.}
		\label{fig:clients}
	\end{figure*}

	We further evaluate the classification accuracy of the FL methods for each category of the client's local data.
	In the non-IID data setting, the number of samples and categories varies widely within each client.
	Figure~\ref{fig:clients} lists two randomly selected clients and shows corresponding classification accuracy for each category.
	The results in Figure~\ref{fig:clients} show that the baseline methods on non-IID data can perform better in head classes while performance drops in predicting tail classes. 
	Our results provide stable performance for each client in classification, both on head and tail classes. 
	
	The comparison between baseline methods indicates that the \texttt{RepPer} outperforms all the other methods in classification accuracy. 
	The results show that \texttt{RepPer} leads to more robust and stable convergence than alternatives on heterogeneous datasets. 
	We attribute this to the fact that the reasonable local representation models improve the robustness and quality of the global model, and \texttt{RepPer} allows the global representation model to be personalized for each specific client.
	
	\begin{figure}[ht]
		\centering
		\includegraphics[width=0.9\columnwidth]{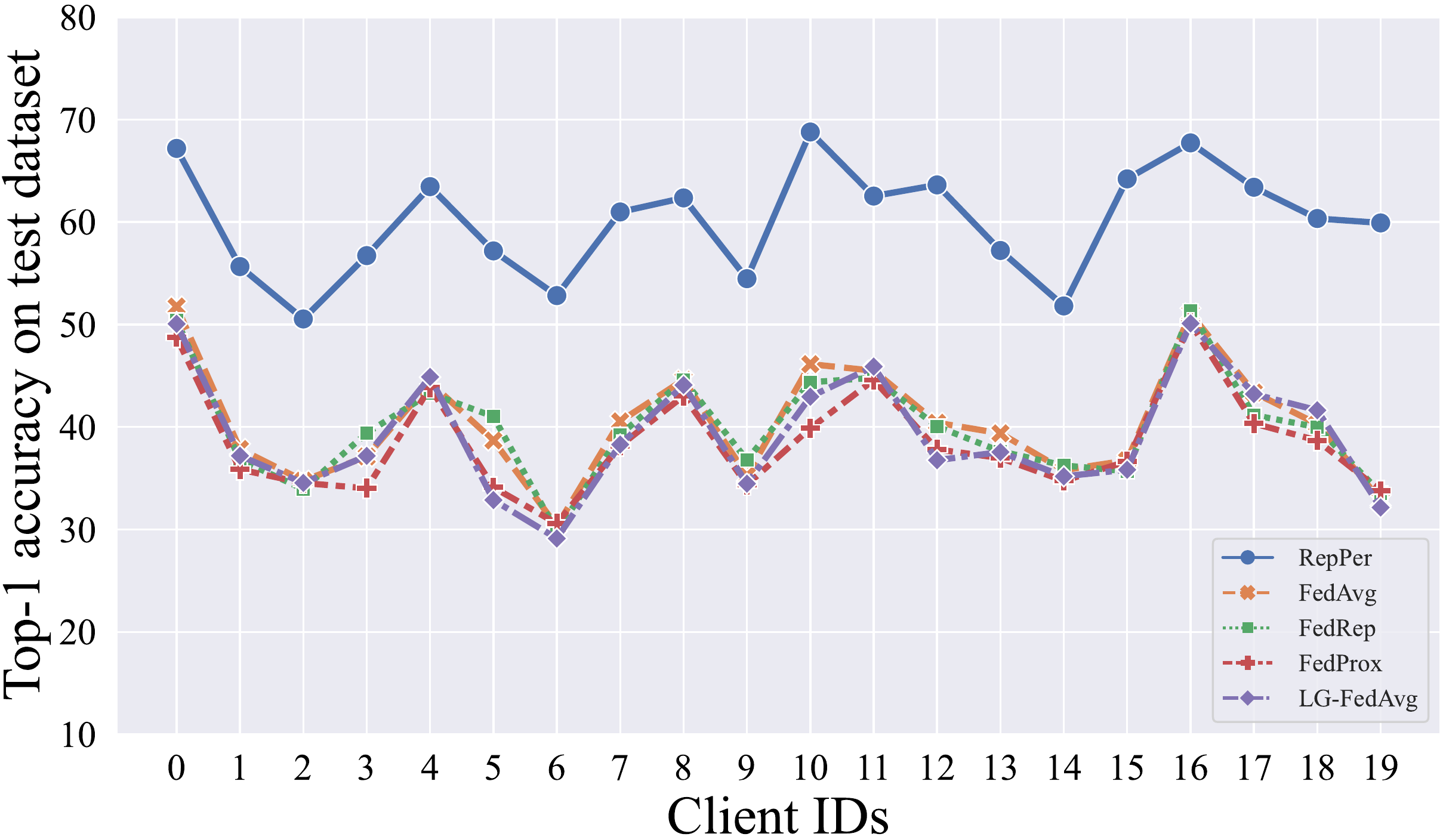}
		\caption{\textbf{Comparing the effect of generalization on new 20 clients} with retraining classifier heads of baseline methods and \texttt{RepPer} on the CIFAR-10 dataset. 
			We consider $20$ new clients with $C=1$ participation ratio and the degree of non-IID with $\alpha=1$. 
			After iterating $100$ times for each personalized classifier head, \texttt{RepPer} can better personalize for all the $20$ new clients compared with alternatives. 
			The increasing number of new clients makes it harder to personalize; however, \texttt{RepPer} outperforms alternatives. }
		\label{fig:new_client}
	\end{figure} 
	
	\subsubsection{Generalization to New Clients}
	We evaluate the strength of \texttt{RepPer} to adapt to new clients. 
	We consider the non-IID federated setting where the resultant global model optimizes for a new client with new target distribution.
	This can be done by performing a few local updates to learn personalized classifier heads from the given global representation model in \texttt{RepPer}.
	
	Accordingly, we train FedAvg, FedProx, LG-FedAvg, FedRep, and \texttt{RepPer}  on the non-IID CIFAR-100 dataset, which contains images of 100 categories. 
	For new clients with local data distribution from the new target domain, which is the CIFAR-10 dataset in the setting of this experiment, we train classifier heads for 100 iterations on its local data while keeping the global representation model parameters fixed.
	
	Figure~\ref{fig:new_client} shows the results of applying ready learned FL methods to new clients that update one-layer MLP as their corresponding classifier heads on the CIFAR-10 dataset. 
	In this experiment, either CIFAR-100 or CIFAR-10 datasets are under heterogeneous settings with the non-IID degree with $\alpha=1$. 
	We show the classification accuracy of the test datasets for 20 new joint clients in Figure~\ref{fig:new_client}. 
	\texttt{RepPer} consistently outperforms alternatives over all the new clients. 
	We attribute the effective personalization for new clients of \texttt{RepPer} due to the semantic discrimination capability of the global representation model, which can be well adapted to a new client with different data distribution.

	\subsection{Flexibility for Low Power Internet-of-Things}  \label{sec:felxibility}
	FL in the Internet of Things (IoT) involves edge devices with diverse hardware and different computation power\cite{IoT1, IoT2,IoT3}. 
	Significantly, some clients with less computation power or even clients that only support classification, can barely participate in FL iterations. 
	To mitigate this problem, we exploit the personalized properties of the \texttt{RepPer} to define various simple classifiers for each client flexibly.
	
	We investigate three machine learning classifiers in the PCL stage of \texttt{RepPer} for personalized prediction: support vector machine (SVM), logistic regression (LR) and multi-layer perceptron (MLP) neural network. 
	For clients with limited computational power, only classifiers need to be retrained by the obtained representations from the global representation model.
	MLP classifiers can be considered a dividing hyperplane in the feature representation space, only learning to make classification decisions.
	Therefore, we retrain these classifiers for clients with limited computation power. 
	Likewise, we retrain linear classifiers (in particular, logistic regression and SVM) for trainable baseline methods personalization with their global representations.	
	
	We compared the performance of \texttt{RepPer} with baseline methods. 
	As shown in Table~\ref{tb:flexible}, \texttt{RepPer} consistently outperform alternatives in personalized classification accuracy.
	We attribute this to a well-learned global representation model and flexibility in machine learning classifier choosing.
	As IoT applications are incredibly diversified, there is a strong need for flexible solutions in the federated learning frameworks. 
	We evaluate that \texttt{RepPer} could be a viable option for low computational power IoT devices.
	
	\begin{table}[ht]
		\caption{{\bf Top-1 accuracy ($\%$) personalized classification with flexible classifiers} by using the trained global model. 
			\texttt{RepPer} outperforms other alternatives across MLP, SVM and logistic regression (LR).
			Best results are marked in \textbf{bold}.}
		
		\label{tb:flexible}
		\begin{center}
			\renewcommand\arraystretch{1.2}
			\scalebox{0.999}{
				\begin{tabular}{*{5}{c}}
					\toprule[1pt]
				Method & MLP & SVM & LR\\
				\midrule[1pt]
				FedAvg         &72.64 &68.66 &68.79\\
				LG-FedAvg           &74.64 &59.82 &60.05\\
				FedRep            &73.03 &70.71 &71.14\\
				{\bf \texttt{RepPer}}            &{\bf 86.22} &{\bf 82.67} &{\bf 83.78}\\
				\bottomrule[1pt]
		\end{tabular}}
	\end{center}
\end{table}

\subsection{Robustness of Different Backbones}
In this section, we conduct and analyze the robustness of the backbones to characterize the \texttt{RepPer}. 
We claim that the CRL stage in \texttt{RepPer} can learn semantic feature representation clusters for each client and consequently lead to a better global model in the federation. 
To validate that it is more robust than alternatives, we extend the comparisons above to the representation model size.

We consider the comparison at different scales of backbones, i.e., ResNet-34 ($\thicksim$~22million parameters) and MobileNet~\cite{mobilenet} ($\thicksim$~3million parameters), which is frequently used in low-power devices.
Accordingly, we train the respective global models with suggested backbones and corresponding personalized classifiers to compare the performance with each other.
The experiment evaluates this two backbones and heterogeneous data with $\alpha= 1$, the fraction of clients $C = 0.4$ for $20$ clients. 
There are $100$ communication rounds and 10 local training epochs per round. 

Table~\ref{tb:sys} visualized the result of FedAvg, LG- FedAvg, FedRep, and \texttt{RepPer} on the CIFAR-10/100 dataset. 
For MobileNet, it shows that \texttt{RepPer}'s personalized model is $17.47\%$ and $3.19\%$ more accurate than alternatives on CIFAR-10 and CIFAR-100, respectively. 
For larger model ResNet-34,  \texttt{RepPer} shows superior classification accuracy by $11.54\%$ and $14.18\%$ than alternatives on CIFAR-10 and CIFAR-100.
We show that the proposed \texttt{RepPer} dominates test classification accuracy in each network architecture as backbones. 
All these results further corroborate that \texttt{RepPer} provides the possibility of mitigating system heterogeneity that alleviates extra effort in altering neural architectures. 
Table~\ref{tb:sys} visualizes \texttt{RepPer} deployed to the portable backbone, such as MobileNet, exhibiting better stability, and higher accuracy than alternatives in personalized FL.

\begin{table}[ht]
	\begin{center}
		\renewcommand\arraystretch{1.4}
		\caption{{\bf Top-1 classification accuracy ($\%$) on the commonly used backbone (ResNet-34) and mobile backbone (MobileNet) on the test set of CIFAR-10/100.} 
			We consider 20 clients for CIFAR-10/100 with $C=0.4$, $\alpha=1$, $E=10$ local epochs per communication round.
			Best results are marked in \textbf{bold}.
		}
		\label{tb:sys}
		\scalebox{0.85}{
				\begin{tabular}{*{10}{c}}
					\toprule[1pt]
					\multirow{2} * {Method} & \multicolumn{2}{c}{\bf CIFAR-10}  & \multicolumn{2}{c}{\bf CIFAR-100} \\
					\cmidrule(lr){2-3} \cmidrule(lr){4-5} 
					& MobileNet &ResNet-34 & MobileNet &ResNet-34 \\
					\midrule[1pt]
					FedAvg         &59.22 &72.64 &33.36   &42.57\\
					LG-FedAvg        &61.63 &74.68 &34.38  &44.67 \\
					FedRep           &60.50 &73.03 &34.48  &40.57\\
					\textbf{\texttt{RepPer}}   &{\bf 79.10} &\textbf{86.22} &{\bf 37.67}  &{\bf 58.85}\\
					\bottomrule[1pt]
			\end{tabular}}
		\end{center}
	\end{table}
	
	\section{Conclusion and future work}
	In this work, we propose \texttt{RepPer} as a personalized FL framework that can adapt to non-IID data to improve the personalized FL performance.
	We divide the traditional FL into two stages that first enable the server to learn a common representation from heterogeneous data in the federation. We then personalize by allowing each client to compute a personalized classifier on their local data using the common representation from stage one. 
	We simulate numbers of non-IID distribution scenarios, where experiments results show our method outperforms previous methods in flexibility and personalization.
	The two-stage training scheme opens questions about the optimal learning scheme compared with end-to-end learning as general training methods. 
	We expected the proposed two-stage personalized federated learning scheme to provide a more flexible paradigm for FL and IoT applications.

\bibliography{aaai22}

\begin{thebibliography}{53}
\providecommand{\natexlab}[1]{#1}

\bibitem[{Arivazhagan et~al.(2019)Arivazhagan, Aggarwal, Singh, and
  Choudhary}]{trans:FedPer}
Arivazhagan, M.~G.; Aggarwal, V.; Singh, A.~K.; and Choudhary, S. 2019.
\newblock Federated learning with personalization layers.
\newblock \emph{arXiv preprint arXiv:1912.00818}.

\bibitem[{Bonawitz et~al.(2017)Bonawitz, Ivanov, Kreuter, Marcedone, McMahan,
  Patel, Ramage, Segal, and Seth}]{feddata-2}
Bonawitz, K.; Ivanov, V.; Kreuter, B.; Marcedone, A.; McMahan, H.~B.; Patel,
  S.; Ramage, D.; Segal, A.; and Seth, K. 2017.
\newblock Practical secure aggregation for privacy-preserving machine learning.
\newblock In \emph{proceedings of the 2017 ACM SIGSAC Conference on Computer
  and Communications Security}, 1175--1191.

\bibitem[{Caldas et~al.(2018)Caldas, Duddu, Wu, Li, Kone{\v{c}}n{\`y}, McMahan,
  Smith, and Talwalkar}]{leaf}
Caldas, S.; Duddu, S. M.~K.; Wu, P.; Li, T.; Kone{\v{c}}n{\`y}, J.; McMahan,
  H.~B.; Smith, V.; and Talwalkar, A. 2018.
\newblock Leaf: A benchmark for federated settings.
\newblock \emph{arXiv preprint arXiv:1812.01097}.

\bibitem[{Chen et~al.(2018)Chen, Luo, Dong, Li, and He}]{meta:fedmeta}
Chen, F.; Luo, M.; Dong, Z.; Li, Z.; and He, X. 2018.
\newblock Federated meta-learning with fast convergence and efficient
  communication.
\newblock \emph{arXiv preprint arXiv:1802.07876}.

\bibitem[{Chen and Chao(2021)}]{bridging}
Chen, H.-Y.; and Chao, W.-L. 2021.
\newblock On Bridging Generic and Personalized Federated Learning for Image
  Classification.
\newblock In \emph{International Conference on Learning Representations}.

\bibitem[{Chen et~al.(2020{\natexlab{a}})Chen, Kornblith, Norouzi, and
  Hinton}]{ssl:simclr}
Chen, T.; Kornblith, S.; Norouzi, M.; and Hinton, G. 2020{\natexlab{a}}.
\newblock A simple framework for contrastive learning of visual
  representations.
\newblock In \emph{International conference on machine learning}, 1597--1607.
  PMLR.

\bibitem[{Chen and He(2021)}]{ssl:simsam}
Chen, X.; and He, K. 2021.
\newblock Exploring simple siamese representation learning.
\newblock In \emph{Proceedings of the IEEE/CVF Conference on Computer Vision
  and Pattern Recognition}, 15750--15758.

\bibitem[{Chen et~al.(2020{\natexlab{b}})Chen, Qin, Wang, Yu, and
  Gao}]{trans:FedHealth}
Chen, Y.; Qin, X.; Wang, J.; Yu, C.; and Gao, W. 2020{\natexlab{b}}.
\newblock Fedhealth: A federated transfer learning framework for wearable
  healthcare.
\newblock \emph{IEEE Intelligent Systems}, 35(4): 83--93.

\bibitem[{Collins et~al.(2021)Collins, Hassani, Mokhtari, and
  Shakkottai}]{Rep:FedRep}
Collins, L.; Hassani, H.; Mokhtari, A.; and Shakkottai, S. 2021.
\newblock Exploiting Shared Representations for Personalized Federated
  Learning.
\newblock \emph{arXiv preprint arXiv:2102.07078}.

\bibitem[{Cortes and Vapnik(1995)}]{svm}
Cortes, C.; and Vapnik, V. 1995.
\newblock Support-vector networks.
\newblock \emph{Machine learning}, 20(3): 273--297.

\bibitem[{Darlow et~al.(2018)Darlow, Crowley, Antoniou, and Storkey}]{cinic}
Darlow, L.~N.; Crowley, E.~J.; Antoniou, A.; and Storkey, A.~J. 2018.
\newblock Cinic-10 is not imagenet or cifar-10.
\newblock \emph{arXiv preprint arXiv:1810.03505}.

\bibitem[{Deng, Kamani, and Mahdavi(2020)}]{RN43}
Deng, Y.; Kamani, M.~M.; and Mahdavi, M. 2020.
\newblock Adaptive personalized federated learning.
\newblock \emph{arXiv preprint arXiv:2003.13461}.
\newblock Client-drift.

\bibitem[{Fallah, Mokhtari, and Ozdaglar(2020)}]{meta:per-fedavg}
Fallah, A.; Mokhtari, A.; and Ozdaglar, A. 2020.
\newblock Personalized federated learning with theoretical guarantees: A
  model-agnostic meta-learning approach.
\newblock \emph{Advances in Neural Information Processing Systems}, 33:
  3557--3568.

\bibitem[{Finn, Abbeel, and Levine(2017)}]{meta:maml}
Finn, C.; Abbeel, P.; and Levine, S. 2017.
\newblock Model-agnostic meta-learning for fast adaptation of deep networks.
\newblock In \emph{International Conference on Machine Learning}, 1126--1135.
  PMLR.

\bibitem[{Grill et~al.(2020)Grill, Strub, Altch{\'e}, Tallec, Richemond,
  Buchatskaya, Doersch, Pires, Guo, Azar et~al.}]{ssl:byol}
Grill, J.-B.; Strub, F.; Altch{\'e}, F.; Tallec, C.; Richemond, P.~H.;
  Buchatskaya, E.; Doersch, C.; Pires, B.~A.; Guo, Z.~D.; Azar, M.~G.; et~al.
  2020.
\newblock Bootstrap your own latent: A new approach to self-supervised
  learning.
\newblock \emph{arXiv preprint arXiv:2006.07733}.

\bibitem[{Gutmann and Hyv{\"a}rinen(2010)}]{ssl:nce}
Gutmann, M.; and Hyv{\"a}rinen, A. 2010.
\newblock Noise-contrastive estimation: A new estimation principle for
  unnormalized statistical models.
\newblock In \emph{Proceedings of the thirteenth international conference on
  artificial intelligence and statistics}, 297--304. JMLR Workshop and
  Conference Proceedings.

\bibitem[{He et~al.(2020{\natexlab{a}})He, Li, So, Zeng, Zhang, Wang, Wang,
  Vepakomma, Singh, Qiu et~al.}]{fedml}
He, C.; Li, S.; So, J.; Zeng, X.; Zhang, M.; Wang, H.; Wang, X.; Vepakomma, P.;
  Singh, A.; Qiu, H.; et~al. 2020{\natexlab{a}}.
\newblock Fedml: A research library and benchmark for federated machine
  learning.
\newblock \emph{arXiv preprint arXiv:2007.13518}.

\bibitem[{He et~al.(2020{\natexlab{b}})He, Fan, Wu, Xie, and
  Girshick}]{ssl:moco}
He, K.; Fan, H.; Wu, Y.; Xie, S.; and Girshick, R. 2020{\natexlab{b}}.
\newblock Momentum contrast for unsupervised visual representation learning.
\newblock In \emph{Proceedings of the IEEE/CVF Conference on Computer Vision
  and Pattern Recognition}, 9729--9738.

\bibitem[{He et~al.(2016)He, Zhang, Ren, and Sun}]{resnet}
He, K.; Zhang, X.; Ren, S.; and Sun, J. 2016.
\newblock Deep residual learning for image recognition.
\newblock In \emph{Proceedings of the IEEE conference on computer vision and
  pattern recognition}, 770--778.

\bibitem[{Howard et~al.(2017)Howard, Zhu, Chen, Kalenichenko, Wang, Weyand,
  Andreetto, and Adam}]{mobilenet}
Howard, A.~G.; Zhu, M.; Chen, B.; Kalenichenko, D.; Wang, W.; Weyand, T.;
  Andreetto, M.; and Adam, H. 2017.
\newblock Mobilenets: Efficient convolutional neural networks for mobile vision
  applications.
\newblock \emph{arXiv preprint arXiv:1704.04861}.

\bibitem[{Hsieh et~al.(2020)Hsieh, Phanishayee, Mutlu, and
  Gibbons}]{divergence-2}
Hsieh, K.; Phanishayee, A.; Mutlu, O.; and Gibbons, P. 2020.
\newblock The non-iid data quagmire of decentralized machine learning.
\newblock In \emph{International Conference on Machine Learning}, 4387--4398.
  PMLR.

\bibitem[{Hsu, Qi, and Brown(2019)}]{dirichlet:2}
Hsu, T.-M.~H.; Qi, H.; and Brown, M. 2019.
\newblock Measuring the effects of non-identical data distribution for
  federated visual classification.
\newblock \emph{arXiv preprint arXiv:1909.06335}.

\bibitem[{Huang et~al.(2021)Huang, Chu, Zhou, Wang, Liu, Pei, and
  Zhang}]{RT:FedAMP}
Huang, Y.; Chu, L.; Zhou, Z.; Wang, L.; Liu, J.; Pei, J.; and Zhang, Y. 2021.
\newblock Personalized Cross-Silo Federated Learning on Non-IID Data.
\newblock \emph{Proceedings of the AAAI Conference on Artificial Intelligence},
  35(9): 7865--7873.

\bibitem[{Kairouz et~al.(2019)Kairouz, McMahan, Avent, Bellet, Bennis, Bhagoji,
  Bonawitz, Charles, Cormode, Cummings et~al.}]{advance}
Kairouz, P.; McMahan, H.~B.; Avent, B.; Bellet, A.; Bennis, M.; Bhagoji, A.~N.;
  Bonawitz, K.; Charles, Z.; Cormode, G.; Cummings, R.; et~al. 2019.
\newblock Advances and open problems in federated learning.
\newblock \emph{arXiv preprint arXiv:1912.04977}.

\bibitem[{Kang et~al.(2020)Kang, Li, Xie, Yuan, and Feng}]{kang2020exploring}
Kang, B.; Li, Y.; Xie, S.; Yuan, Z.; and Feng, J. 2020.
\newblock Exploring balanced feature spaces for representation learning.
\newblock In \emph{International Conference on Learning Representations}.

\bibitem[{Karimireddy et~al.(2020)Karimireddy, Kale, Mohri, Reddi, Stich, and
  Suresh}]{scaffold}
Karimireddy, S.~P.; Kale, S.; Mohri, M.; Reddi, S.; Stich, S.; and Suresh,
  A.~T. 2020.
\newblock Scaffold: Stochastic controlled averaging for federated learning.
\newblock In \emph{International Conference on Machine Learning}, 5132--5143.
  PMLR.

\bibitem[{Khodak, Balcan, and Talwalkar(2019)}]{meta:aruba}
Khodak, M.; Balcan, M.; and Talwalkar, A. 2019.
\newblock Adaptive Gradient-Based Meta-Learning Methods.
\newblock In \emph{Neural Information Processing Systems}.

\bibitem[{Khosla et~al.(2020)Khosla, Teterwak, Wang, Sarna, Tian, Isola,
  Maschinot, Liu, and Krishnan}]{supcon}
Khosla, P.; Teterwak, P.; Wang, C.; Sarna, A.; Tian, Y.; Isola, P.; Maschinot,
  A.; Liu, C.; and Krishnan, D. 2020.
\newblock Supervised contrastive learning.
\newblock \emph{arXiv preprint arXiv:2004.11362}.

\bibitem[{Kingma and Ba(2014)}]{adam}
Kingma, D.~P.; and Ba, J. 2014.
\newblock Adam: A method for stochastic optimization.
\newblock \emph{arXiv preprint arXiv:1412.6980}.

\bibitem[{Kone{\v{c}}n{\`y} et~al.(2016)Kone{\v{c}}n{\`y}, McMahan, Yu,
  Richt{\'a}rik, Suresh, and Bacon}]{feddata-1}
Kone{\v{c}}n{\`y}, J.; McMahan, H.~B.; Yu, F.~X.; Richt{\'a}rik, P.; Suresh,
  A.~T.; and Bacon, D. 2016.
\newblock Federated learning: Strategies for improving communication
  efficiency.
\newblock \emph{arXiv preprint arXiv:1610.05492}.

\bibitem[{Krizhevsky(2012)}]{cifar}
Krizhevsky, A. 2012.
\newblock Learning Multiple Layers of Features from Tiny Images.
\newblock \emph{University of Toronto}.

\bibitem[{Li et~al.(2020)Li, Sahu, Talwalkar, and Smith}]{challenge}
Li, T.; Sahu, A.~K.; Talwalkar, A.; and Smith, V. 2020.
\newblock Federated learning: Challenges, methods, and future directions.
\newblock \emph{IEEE Signal Processing Magazine}, 37(3): 50--60.

\bibitem[{Li et~al.(2018)Li, Sahu, Zaheer, Sanjabi, Talwalkar, and
  Smith}]{FRT:FedProx}
Li, T.; Sahu, A.~K.; Zaheer, M.; Sanjabi, M.; Talwalkar, A.; and Smith, V.
  2018.
\newblock Federated optimization in heterogeneous networks.
\newblock \emph{arXiv preprint arXiv:1812.06127}.

\bibitem[{Liang et~al.(2020)Liang, Liu, Ziyin, Allen, Auerbach, Brent,
  Salakhutdinov, and Morency}]{Rep:LG-Fedavg}
Liang, P.~P.; Liu, T.; Ziyin, L.; Allen, N.~B.; Auerbach, R.~P.; Brent, D.;
  Salakhutdinov, R.; and Morency, L.-P. 2020.
\newblock Think locally, act globally: Federated learning with local and global
  representations.
\newblock \emph{arXiv preprint arXiv:2001.01523}.

\bibitem[{Lin et~al.(2020)Lin, Kong, Stich, and Jaggi}]{FedDF}
Lin, T.; Kong, L.; Stich, S.~U.; and Jaggi, M. 2020.
\newblock Ensemble distillation for robust model fusion in federated learning.
\newblock \emph{arXiv preprint arXiv:2006.07242}.

\bibitem[{Luo et~al.(2021)Luo, Chen, Hu, Zhang, Liang, and Feng}]{non-iid:cls}
Luo, M.; Chen, F.; Hu, D.; Zhang, Y.; Liang, J.; and Feng, J. 2021.
\newblock No fear of heterogeneity: Classifier calibration for federated
  learning with non-iid data.
\newblock \emph{Advances in Neural Information Processing Systems}, 34.

\bibitem[{McMahan et~al.(2017)McMahan, Moore, Ramage, Hampson, and
  y~Arcas}]{fedavg}
McMahan, B.; Moore, E.; Ramage, D.; Hampson, S.; and y~Arcas, B.~A. 2017.
\newblock Communication-efficient learning of deep networks from decentralized
  data.
\newblock In \emph{Artificial intelligence and statistics}, 1273--1282. PMLR.

\bibitem[{Min et~al.(2019)Min, Xiao, Chen, Cheng, Wu, and Zhuang}]{IoT2}
Min, M.; Xiao, L.; Chen, Y.; Cheng, P.; Wu, D.; and Zhuang, W. 2019.
\newblock Learning-based computation offloading for IoT devices with energy
  harvesting.
\newblock \emph{IEEE Transactions on Vehicular Technology}, 68(2): 1930--1941.

\bibitem[{Mohri, Sivek, and Suresh(2019)}]{feddata-3}
Mohri, M.; Sivek, G.; and Suresh, A.~T. 2019.
\newblock Agnostic federated learning.
\newblock In \emph{International Conference on Machine Learning}, 4615--4625.
  PMLR.

\bibitem[{Ning et~al.(2018)Ning, Dong, Kong, and Xia}]{IoT1}
Ning, Z.; Dong, P.; Kong, X.; and Xia, F. 2018.
\newblock A cooperative partial computation offloading scheme for mobile edge
  computing enabled Internet of Things.
\newblock \emph{IEEE Internet of Things Journal}, 6(3): 4804--4814.

\bibitem[{Oh, Kim, and Yun(2021)}]{fedbabu}
Oh, J.; Kim, S.; and Yun, S.-Y. 2021.
\newblock FedBABU: Towards Enhanced Representation for Federated Image
  Classification.
\newblock \emph{arXiv preprint arXiv:2106.06042}.

\bibitem[{Pang et~al.(2020)Pang, Huang, Xie, Han, and Cai}]{IoT3}
Pang, J.; Huang, Y.; Xie, Z.; Han, Q.; and Cai, Z. 2020.
\newblock Realizing the heterogeneity: a self-organized federated learning
  framework for IoT.
\newblock \emph{IEEE Internet of Things Journal}, 8(5): 3088--3098.

\bibitem[{T~Dinh, Tran, and Nguyen(2020)}]{RT:pFedMe}
T~Dinh, C.; Tran, N.; and Nguyen, T.~D. 2020.
\newblock Personalized Federated Learning with Moreau Envelopes.
\newblock \emph{Advances in Neural Information Processing Systems}, 33.

\bibitem[{Tian, Krishnan, and Isola(2020)}]{ssl:cmc}
Tian, Y.; Krishnan, D.; and Isola, P. 2020.
\newblock Contrastive multiview coding.
\newblock In \emph{Computer Vision--ECCV 2020: 16th European Conference,
  Glasgow, UK, August 23--28, 2020, Proceedings, Part XI 16}, 776--794.
  Springer.

\bibitem[{Van~der Maaten and Hinton(2008)}]{tsne}
Van~der Maaten, L.; and Hinton, G. 2008.
\newblock Visualizing data using t-SNE.
\newblock \emph{Journal of machine learning research}, 9(11).

\bibitem[{Wang et~al.(2020)Wang, Yurochkin, Sun, Papailiopoulos, and
  Khazaeni}]{RN20}
Wang, H.; Yurochkin, M.; Sun, Y.; Papailiopoulos, D.; and Khazaeni, Y. 2020.
\newblock Federated learning with matched averaging.
\newblock \emph{arXiv preprint arXiv:2002.06440}.

\bibitem[{Wang et~al.(2021)Wang, Han, Wei, Zhang, and
  Wang}]{wang2021contrastive}
Wang, P.; Han, K.; Wei, X.-S.; Zhang, L.; and Wang, L. 2021.
\newblock Contrastive learning based hybrid networks for long-tailed image
  classification.
\newblock In \emph{Proceedings of the IEEE/CVF Conference on Computer Vision
  and Pattern Recognition}, 943--952.

\bibitem[{Yurochkin et~al.(2019)Yurochkin, Agarwal, Ghosh, Greenewald, Hoang,
  and Khazaeni}]{dirichlet:1}
Yurochkin, M.; Agarwal, M.; Ghosh, S.; Greenewald, K.; Hoang, N.; and Khazaeni,
  Y. 2019.
\newblock Bayesian nonparametric federated learning of neural networks.
\newblock In \emph{International Conference on Machine Learning}, 7252--7261.
  PMLR.

\bibitem[{Zhang et~al.(2021{\natexlab{a}})Zhang, Xie, Bai, Yu, Li, and
  Gao}]{survey1}
Zhang, C.; Xie, Y.; Bai, H.; Yu, B.; Li, W.; and Gao, Y. 2021{\natexlab{a}}.
\newblock A survey on federated learning.
\newblock \emph{Knowledge-Based Systems}, 216: 106775.

\bibitem[{Zhang et~al.(2020)Zhang, Kuang, You, Shen, Xiao, Zhang, Wu, Zhuang,
  and Li}]{Rep:FedCA}
Zhang, F.; Kuang, K.; You, Z.; Shen, T.; Xiao, J.; Zhang, Y.; Wu, C.; Zhuang,
  Y.; and Li, X. 2020.
\newblock Federated unsupervised representation learning.
\newblock \emph{arXiv preprint arXiv:2010.08982}.

\bibitem[{Zhang et~al.(2021{\natexlab{b}})Zhang, Li, Ma, Luo, and
  Li}]{zhang2021federated}
Zhang, W.; Li, X.; Ma, H.; Luo, Z.; and Li, X. 2021{\natexlab{b}}.
\newblock Federated learning for machinery fault diagnosis with dynamic
  validation and self-supervision.
\newblock \emph{Knowledge-Based Systems}, 213: 106679.

\bibitem[{Zhao et~al.(2018)Zhao, Li, Lai, Suda, Civin, and
  Chandra}]{divergence-1}
Zhao, Y.; Li, M.; Lai, L.; Suda, N.; Civin, D.; and Chandra, V. 2018.
\newblock Federated learning with non-iid data.
\newblock \emph{arXiv preprint arXiv:1806.00582}.

\bibitem[{Zhuang et~al.(2021)Zhuang, Gan, Wen, Zhang, and Yi}]{Rep:BYOL}
Zhuang, W.; Gan, X.; Wen, Y.; Zhang, S.; and Yi, S. 2021.
\newblock Collaborative Unsupervised Visual Representation Learning from
  Decentralized Data.
\newblock In \emph{Proceedings of the IEEE/CVF International Conference on
  Computer Vision}, 4912--4921.

\end{thebibliography}
\newpage
\ \\
\newpage
\appendix
\section{Gradient Derivation}	
\label{A1}
In Section 5, we claim that SC loss helps \texttt{RepPer} to learn local representations on the non-IID data. 
In this section, we perform gradient derivations with respect to normalized representation $z$ and feature representation before normalization $r$, proving an intrinsic property of SC loss that works well for learning representations on non-IID data.
We start by deriving the gradient with respect to $z$ of Eq.~\eqref{loss:CLR2}:
\begin{equation}
	\begin{split}
		& \frac{\partial \ell_j}{\partial z_{j}}\\
		=& -\frac{\partial}{\partial z_{j}}    \log \left\{
		\frac{1}{|P(j)|} 
		\sum_{p \in P(j)} \frac{e^ {\left(z_{j} \cdot z_{p} / \tau\right)}}
		{\sum_{a \in A(j)} e^ {\left(z_{j} \cdot z_{a} / \tau\right)}}
		\right\} \\
		=& \frac{\partial}{\partial z_{j}} \log \sum_{a \in A(j)} e^{\left(z_{j} \cdot z_{a} / \tau\right)}
		-\frac{\partial}{\partial z_{j}} \log \sum_{p \in P(j)} e^{ \left(z_{j} \cdot z_{p} / \tau\right)} \\
		=& \frac{1}{\tau} 
		\frac{\sum_{a \in A(j)} z_{a} e^{(z_{j} \cdot z_{a} / \tau)}}
		{\sum_{a \in A(j)} e^{\left(z_{j} \cdot z_{a} / \tau\right)}}
		-\frac{1}{\tau} 
		\frac{\sum_{p \in P(j)} z_{p} e^{\left(z_{j} \cdot z_{p} / \tau\right)}}
		{\sum_{p \in P(j)} e^{z_{j} \cdot z_{p} / \tau)}}  \\
		=&\frac{1}{\tau} 
		\frac{\sum_{p \in P(j)} z_{p} e^{ \left(z_{j} \cdot z_{p} / \tau\right)}
			+\sum_{n \in N(j)} z_{n} e^{ \left(z_{j} \cdot z_{n} / \tau\right)}}
		{\sum_{a \in A(j)} e^{\left(z_{j} \cdot z_{a} / \tau\right)}}\\
		& \quad -\frac{1}{\tau} 
		\frac{\sum_{p \in P(j)} z_{p} e^{\left(z_{j} \cdot z_{p} / \tau\right)}}
		{\sum_{p \in P(j)} e^{ \left(z_{j} \cdot z_{p} / \tau\right)}} \\
		=&\frac{1}{\tau}
		\left\{\sum_{p \in P(j)} z_{p}\left(P_{jp} - X_{jp}\right)+\sum_{n \in N(j)} z_{n} P_{jn}\right\}, 
		\label{eq:gradz}
	\end{split}
\end{equation}
where
\begin{equation}
	P_{j p} = \frac{e^{\left(z_{j} \cdot z_{p} / \tau\right)}}
	{\sum_{a \in A(j)} e^{\left(z_{j} \cdot z_{a} / \tau\right)}},
\end{equation}
\begin{equation}
	X_{j p} = \frac{e^{\left(z_{j} \cdot z_{p} / \tau\right)}}
	{\sum_{p^{\prime} \in P(j)} e{\left(z_{j} \cdot z_{p^{\prime}} / \tau\right)}}.
\end{equation}

In local representation learning, the normalized representation $z$ and representation before normalized $r$ have the relation: $z_j=r_j/\|r_j\|$.
Therefore, the gradient of the SC loss with respect to $r$ is related to that with respect to $z$ via the chain rule:
\begin{equation}
	\frac{\partial \ell_j(z_j)}{\partial r_j}
	=\frac{\partial \ell_j(z_j)}{\partial z_j}
	\cdot \frac{\partial z_j}{\partial r_j},
\end{equation}
where
\begin{equation}
	\begin{split}
		\frac{\partial z_j}{\partial r_j}
		= &	\frac{\partial }{\partial r_j}
		\left(\frac{r_j}{\parallel r_j \parallel}\right) \\
		=&	\frac{1}{\parallel r_j \parallel} \mathrm{I} 
		- r_j \left(\frac{\partial (1/\parallel r_j\parallel)}
		{\partial r_j}\right)^{T}\\
		=&	\frac{1}{\parallel r_j \parallel}
		\left(\mathrm{I} - \frac{r_j \cdot r_j^{T}}
		{\parallel r_j \parallel ^2}\right)\\
		= & 	\frac{1}{\parallel r_j \parallel} 
		(\mathrm{I} - z_j \cdot z_j^{T}),
		\label{eq:gradzr}
	\end{split}
\end{equation}

We combine Eq.~\eqref{eq:gradz} and \eqref{eq:gradzr} thus give the following:
\begin{equation}
	\begin{split}
		&\frac{\partial \ell_{j}}{\partial r_{j}} \\
		=&	\frac{1}{\tau \parallel r_{j} \parallel}
		\left(\mathrm{I} - z_{j} z_{j}^{T}\right)
		\left\{
		\sum_{p \in P(j)} z_{p}\left(P_{j p} - X_{j p}\right) \right. \\
		&\left. \quad		+\sum_{n \in N(j)} z_{n} P_{j n}
		\right\} \\
		=&	\frac{1}{\tau \parallel r_{j} \parallel}
		\left\{ 
		\sum_{p \in P(j)}\left(z_{p} 
		- \left(z_{j} \cdot z_{p}\right) z_{j}\right)
		\left(P_{j p} - X_{j p}\right) \right.\\
		&\left. \quad + \sum_{n \in N(j)} \left(z_{n} 
		- \left(z_{j} \cdot z_{n}\right) z_{j}\right) P_{jn} 
		\right\} \\
		=&	\left.\frac{\partial \ell_{j}}
		{\partial z_{j}}\right|_{\mathrm{P}(\mathrm{j})}
		+\left.\frac{\partial \ell_{j}}
		{\partial z_{j}}\right|_{\mathrm{N}(\mathrm{j})},
	\end{split}
\end{equation}
where
\begin{flalign}
	&\ \!\!\frac{\partial \ell_{j}}
	{\partial z_{j}}\Bigg|_{\mathrm{P}(\mathrm{j})}
	\!\!=\frac{1}{\tau \parallel \!\! r_{j} \!\! \parallel}
	\!\! \sum_{p \in P(j)}\!\!\! (z_{p} \!-\! (z_{j}\! \cdot \! z_{p}) z_{j})  (P_{j p} \! -  \!X_{j p}), &\\
	&\ \frac{\partial \ell_{j}}
	{\partial z_{j}}\Bigg|_{\mathrm{N}(\mathrm{j})}
	\!\!=\sum_{n \in N(j)} \left(z_{n} 
	- (z_{j} \cdot z_{n}\right) z_{j}) P_{jn}. &
\end{flalign}

\end{document}